\documentclass[letterpaper, 10 pt, conference]{ieeeconf}  

\IEEEoverridecommandlockouts                              

\newcommand\copyrighttext{\footnotesize \textcopyright~2023 IEEE. Personal use of this material is permitted.  Permission from IEEE must be obtained for all other uses, in any current or future media, including reprinting/republishing this material for advertising or promotional purposes, creating new collective works, for resale or redistribution to servers or lists, or reuse of any copyrighted component of this work in other works.%
}
\newcommand\copyrightnotice{%
    \begin{tikzpicture}[remember picture,overlay]%
 	\node[%
        anchor=south, %
        yshift=10pt%
    ] at (current page.south)%
 	{\fbox{\parbox{\dimexpr\textwidth-\fboxsep-\fboxrule\relax}{\copyrighttext}}};%
 	\end{tikzpicture}%
}

\usepackage{graphicx} 
\usepackage{pgfplots}
\usepackage{amsmath} 
\usepackage{multirow}
\usepackage{svg}
\usepackage{hyperref}
\usepackage{siunitx}
\usepackage{booktabs}
\usepackage[section]{placeins}

\title{\LARGE \bf
Efficient Path Planning in Large Unknown Environments with Switchable System Models for Automated Vehicles%
}

\author{Oliver Schumann, Michael Buchholz, and Klaus Dietmayer
\thanks{Part of this work were supported by the State Ministry of Economic Affairs Baden-Württemberg (Project U-Shift II, AZ 3-433.62-DLR/60).}
\thanks{The authors are with the Institute of Measurement, Control and Microtechnology, Ulm University, Albert-Einstein-Allee 41, 89081 Ulm, Germany E-Mail: {\tt\small firstname.lastname@uni-ulm.de}}%
}

\begin{document}

\maketitle
\copyrightnotice%
\thispagestyle{empty}
\pagestyle{empty}

\begin{abstract}
Large environments are challenging for path planning algorithms as the size of the configuration space increases. Furthermore, if the environment is mainly unexplored, large amounts of the path are planned through unknown areas. Hence, a complete replanning of the entire path occurs whenever the path collides with newly discovered obstacles. 
We propose a novel method that stops the path planning algorithm after a certain distance. It is used to navigate the algorithm in large environments and is not prone to problems of existing navigation approaches. Furthermore, we developed a method to detect significant environment changes to allow a more efficient replanning. At last, we extend the path planner to be used in the U-Shift concept vehicle. It can switch to another system model and rotate around the center of its rear axis. The results show that the proposed methods generate nearly identical paths compared to the standard Hybrid A* while drastically reducing the execution time. Furthermore, we show that the extended path planning algorithm enables the efficient use of the maneuvering capabilities of the concept vehicle to plan concise paths in narrow environments.
\end{abstract}

\section{INTRODUCTION}
Large environments pose problems to path planning algorithms of autonomous vehicles due to the large size of the configuration space. It grows exponentially with the size of the environment \cite{bellmann} which results in long runtimes for each planned path. Furthermore, unexplored environments are even more challenging as the vehicle can only gradually explore the environment with its sensors while trying to reach the desired goal. Hence, the majority of obstacles that lie between the vehicle and its goal are not discovered yet and the path is prone to collide with these undetected obstacles in the future. If so, the path is not valid anymore. It must be discarded and a new path is planned. Hence, in unknown regions, most parts of a path are planned unnecessarily which is not an efficient use of energy and processing power. 
Furthermore, the majority of a path up to the collision with a previously discovered obstacle is usually still valid. The recalculation of this path is hence, also not always necessary. In addition to that, even in static environments successively planned paths are not necessarily equal. Varying paths right in front of the vehicle can hence lead to high control inputs of the underlying control algorithm which reduces the smoothness of the effective path. 

\begin{figure}[t]
\vspace{2mm}
	\centering
	\def\svgwidth{1.0\linewidth}
	\graphicspath{{img}}
\begingroup%
  \makeatletter%
  \providecommand\color[2][]{%
    \errmessage{(Inkscape) Color is used for the text in Inkscape, but the package 'color.sty' is not loaded}%
    \renewcommand\color[2][]{}%
  }%
  \providecommand\transparent[1]{%
    \errmessage{(Inkscape) Transparency is used (non-zero) for the text in Inkscape, but the package 'transparent.sty' is not loaded}%
    \renewcommand\transparent[1]{}%
  }%
  \providecommand\rotatebox[2]{#2}%
  \newcommand*\fsize{\dimexpr\f@size pt\relax}%
  \newcommand*\lineheight[1]{\fontsize{\fsize}{#1\fsize}\selectfont}%
  \ifx\svgwidth\undefined%
    \setlength{\unitlength}{413.85826772bp}%
    \ifx\svgscale\undefined%
      \relax%
    \else%
      \setlength{\unitlength}{\unitlength * \real{\svgscale}}%
    \fi%
  \else%
    \setlength{\unitlength}{\svgwidth}%
  \fi%
  \global\let\svgwidth\undefined%
  \global\let\svgscale\undefined%
  \makeatother%
  \begin{picture}(1,0.69178082)%
    \lineheight{1}%
    \setlength\tabcolsep{0pt}%
    \put(0,0){\includegraphics[width=\unitlength,page=1]{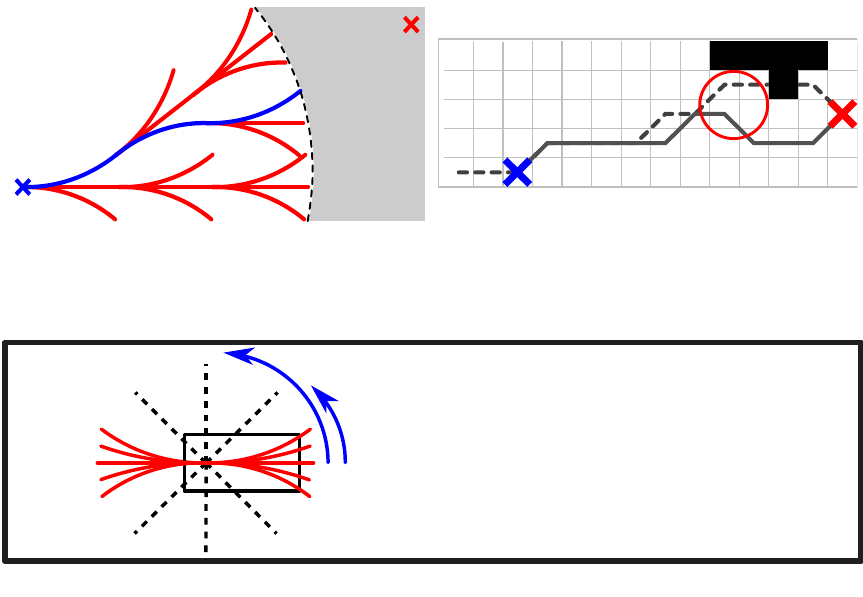}}%
    \put(0.00978034,0.39377121){\makebox(0,0)[lt]{\lineheight{1.25}\smash{\begin{tabular}[t]{l}a)\end{tabular}}}}%
    \put(0.51527441,0.39278192){\makebox(0,0)[lt]{\lineheight{1.25}\smash{\begin{tabular}[t]{l}b)\end{tabular}}}}%
    \put(0.0102689,0.00016194){\makebox(0,0)[lt]{\lineheight{1.25}\smash{\begin{tabular}[t]{l}c)\\\end{tabular}}}}%
    \put(0.80519809,0.35808772){\makebox(0,0)[lt]{\lineheight{1.25}\smash{\begin{tabular}[t]{l}$P\textsubscript{s}$\\\\\end{tabular}}}}%
    \put(0,0){\includegraphics[width=\unitlength,page=2]{drawing_3.pdf}}%
  \end{picture}%
\endgroup%

	\caption{Contributions of this paper: a) The \textit{early stopping} method stops the path planning algorithm to plan paths more efficiently. b) The replanning scheme defines the start pose. c) A path planning algorithm that switches between two system models to use the maneuvering capabilities of the U-Shift~II concept vehicle \cite{ushift1, ushift2}.}
	\label{fig:system}
 \vspace{-3mm}
\end{figure}

These are challenges that are faced during the development of the U-Shift concept vehicle \cite{ushift1, ushift2} which is shown in Fig.~\ref{fig:system}~c). It is a modular transport system that consists of a so-called driveboard for driverless automated transport of different kinds of capsules. Here, the concept vehicle is shown with a capsule to transport passengers. The main difference to other vehicles in the context of trajectory planning is the extended maneuverability of the driveboard. It can move like a normal vehicle and can thus be modeled by a bicycle model during path planning. However, it can also switch to a different model by turning its front wheels to the inside. Then it can rotate around the center point of its rear axis. There exist similar vehicle concepts with extended maneuvering capabilities like the UNICAR\textit{agil} vehicles \cite{unicaragil}. However, to the best of our knowledge, no path planning approaches have been published that can inherently handle the capabilities of these concepts.
Hence, the main contributions of this paper are visualized in Fig.~\ref{fig:system}:
\begin{itemize}
\item A novel navigation method called \textit{early stopping} improves the efficiency of path planning algorithms in large and unknown environments.
\item A replanning scheme based on two-dimensional A*-paths that detects significant environment changes; and
\item Two extensions to the Hybrid A* (HA*) algorithm \cite{hybridastar} which allow the path planning for the U-Shift concept vehicle.

\end{itemize}
The open source code of this project will be published on the final submission of this paper~\footnote{\href{https://github.com/oliver-schumann/guided-extended-hybrid-astar}{https://github.com/oliver-schumann/guided-extended-hybrid-astar}}. 
\newpage
The following part of the paper is structured in the following way. Sec.~\ref{sec:related_work} discusses methods related to path planning in large and unknown environments followed by Sec.~\ref{subsec:nav} in which two methods are proposed to improve path planning in large and unexplored environments. Next, Sec.~\ref{subsec:ra} proposes a method to plan paths for the U-Shift concept vehicle. Then, Sec.~\ref{sec:guidance} \ref{subsec:ra_eval} evaluates the application of stated methods qualitatively on real-world data and investigates them quantitatively on simulated data. The paper ends with Sec.~\ref{sec:conclusion} which summarizes the results.
\section{RELATED WORK}\label{sec:related_work}
One method to simplify and hence speed up path planning tasks in large environments is to extract so-called sub-goals or waypoints from the environment. They serve as the goal for the path planning algorithm if the goal pose is too far away from the start pose. This can be done by applying methods that have a lower dimension and can therefore handle large environments more easily \cite{bellmann}. Common methods use visibility graphs \cite{way1, way2} or two-stage approaches \cite{twostage_hybrid, two-stage-rrt, global_local}. The first ones extract waypoints from the nodes of a visibility graph, the two-stage approaches execute a two-dimensional graph search first and extract waypoints from this search result. A disadvantage of this class of methods is that the resulting path using these navigation approaches may not find the optimal way with respect to the vehicle dynamics as the guiding method uses only a lower-dimensional representation of the environment.

Additionally, previously undetected obstacles can result in sub-optimal paths with respect to the clearance to obstacles if they appear close to the path. This problem can naively be addressed by periodic replanning from the start. However, in most cases, this is a waste of energy and processing power as replanning is not necessary in most cases. Hence, there exist several methods to solve this issue more efficiently.
\cite{rrt_fn} and \cite{rrt_fn_dynamic} propose methods in the field of sampling-based algorithms like rapidly exploring random trees that try to achieve the goal of rewiring the graph on environment changes. 
However, these approaches can not be adapted to the graph-based approach used in this paper, which is required for the guiding method that will be presented.

In grid-based environments, the following algorithms were developed to allow a more efficient replanning: the D* algorithm \cite{dstar}, the D* lite \cite{dstarlite}, and the anytime dynamic A* method \cite{anytime_dynamic_astar}. They can reuse the previously generated nodes and attempt to repair the created graph if the environment changes. However, extensions of these graph-based algorithms for non-holonomic vehicle dynamics, like the Hybrid A* algorithm \cite{hybridastar}, have not been integrated into these algorithms. There exist methods that try to reuse the created nodes up to a collision by pruning all colliding parts of the graph as in \cite{prune}. However, collision-free nodes of the graph behind a colliding node are discarded anyway. 

Hence, a method is needed to efficiently calculate paths in large environments without the necessity to replan periodically and entirely from the beginning if obstacles are discovered.

\section{METHOD} \label{sec:method}
This section proposes a novel navigation method called \textit{early stopping} that guides graph-based path planning algorithms in large environments. Then, a method is presented to allow a more efficient replanning. At last, a method is proposed to integrate switchable system models into the path planning algorithm.
\subsection{Navigation Method} \label{subsec:nav}
At first, a similar method compared to the two-stage method proposed in \cite{twostage_hybrid} is introduced that will serve as a baseline. It can generate a waypoint 
from a two-dimensional A*-path connecting the start and goal positions. The standard Hybrid A* algorithm already requires the calculation of a distance heuristic map. It can be calculated efficiently by a two-dimensional A* algorithm as done in \cite{kurzer} and \cite{ara}. Hence, this distance heuristic can be used to extract an A*-path connecting the start and goal positions. This path is used to sample a waypoint at a certain distance $s\textsubscript{w}$ from the vehicle along this path. The waypoint can be extended to a waypose
by calculating the orientation from two consecutive points. This pose is now used as a goal for the planning algorithm. The only difference to \cite{twostage_hybrid} is that two stags are not explicitly separated but the existing data structures are reused. This method is called \textit{waypoint navigation} in the following. 

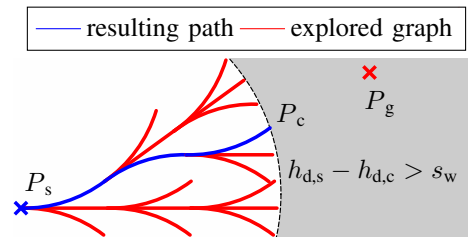
\begin{figure}[tbp]
\vspace{1mm}
	\centering
\begin{minipage}{0.7\columnwidth}
\raggedleft
\begin{tikzpicture}
\begin{axis}[axis line style={draw=none}, tick style={draw=none}, yticklabels={,,}, xticklabels={,,}, width=3cm, height=1.7cm, legend columns=3]
\addplot[color=red, forget plot]{exp(x)};
\pgfplotsset{my area legend/.style={
  legend image code/.code={\fill[#1] (0cm,-0.2cm) rectangle (0.35cm,0.2cm);}
  }
}
\addlegendimage{blue}
\addlegendentry{resulting path}
\addlegendimage{red}
\addlegendentry{explored graph}
\end{axis}
\end{tikzpicture}
\end{minipage}\\
\vspace{1mm}
	\def\svgwidth{0.7\linewidth}
	\graphicspath{{img}}
\begingroup%
  \makeatletter%
  \providecommand\color[2][]{%
    \errmessage{(Inkscape) Color is used for the text in Inkscape, but the package 'color.sty' is not loaded}%
    \renewcommand\color[2][]{}%
  }%
  \providecommand\transparent[1]{%
    \errmessage{(Inkscape) Transparency is used (non-zero) for the text in Inkscape, but the package 'transparent.sty' is not loaded}%
    \renewcommand\transparent[1]{}%
  }%
  \providecommand\rotatebox[2]{#2}%
  \newcommand*\fsize{\dimexpr\f@size pt\relax}%
  \newcommand*\lineheight[1]{\fontsize{\fsize}{#1\fsize}\selectfont}%
  \ifx\svgwidth\undefined%
    \setlength{\unitlength}{552.75590551bp}%
    \ifx\svgscale\undefined%
      \relax%
    \else%
      \setlength{\unitlength}{\unitlength * \real{\svgscale}}%
    \fi%
  \else%
    \setlength{\unitlength}{\svgwidth}%
  \fi%
  \global\let\svgwidth\undefined%
  \global\let\svgscale\undefined%
  \makeatother%
  \begin{picture}(1,0.39285051)%
    \lineheight{1}%
    \setlength\tabcolsep{0pt}%
    \put(0,0){\includegraphics[width=\unitlength,page=1]{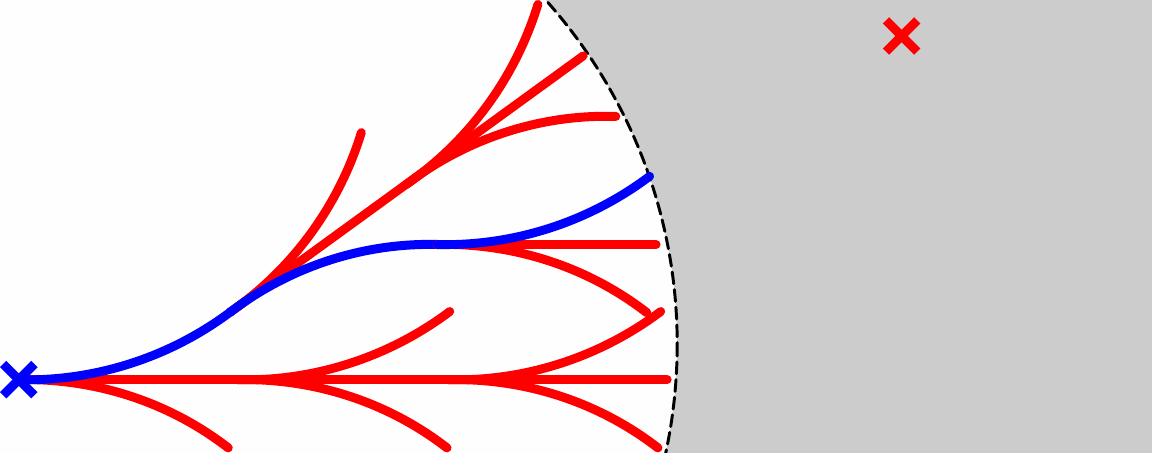}}%
    \put(0.77160069,0.28082124){\makebox(0,0)[lt]{\lineheight{1.25}\smash{\begin{tabular}[t]{l}$P\textsubscript{g}$\\\end{tabular}}}}%
    \put(0.02375347,0.10164033){\makebox(0,0)[lt]{\lineheight{1.25}\smash{\begin{tabular}[t]{l}$P\textsubscript{s}$\\\end{tabular}}}}%
    \put(0.60238979,0.12977781){\makebox(0,0)[lt]{\lineheight{1.25}\smash{\begin{tabular}[t]{l}$h\textsubscript{d,s} - h\textsubscript{d,c} > s\textsubscript{w}$\end{tabular}}}}%
    \put(0.5774725,0.25261982){\makebox(0,0)[lt]{\lineheight{1.25}\smash{\begin{tabular}[t]{l}$P\textsubscript{c}$\\\end{tabular}}}}%
  \end{picture}%
\endgroup%

	\caption{Schematic visualization of the \textit{early stopping} method: The start node is shown as a blue cross, while the goal node is shown as a red one. In the gray area, the distance heuristic decreased for at least $s\textsubscript{w}$.}
	\label{fig:early_stopping}
\end{figure}

Now, a novel navigation method is proposed that is called \textit{early stopping}. It is similar to the first method in that the path planner only plans a path up to a certain distance $s\textsubscript{w}$. However, in this approach, this is done implicitly by stopping the graph search of the Hybrid A* algorithm. The method is shown in Fig.~\ref{fig:early_stopping}. This is done by evaluating the already calculated distance heuristic. If the difference of the distance heuristic of the start node $h\textsubscript{d,s}$ compared to the heuristic of the current node $h\textsubscript{d,c}$ is higher than the distance to plan $s\textsubscript{w}$, the graph search is stopped as denoted in
\begin{equation}
    h\textsubscript{d,s} - h\textsubscript{d,c} > s\textsubscript{w}.
\end{equation}
Hence, there is no explicit pose to plan to but a desired state to be in. 
As the vehicle progresses toward the final goal, it approaches the end of its path. In both methods, every progressed $s\textsubscript{t}$ the planner either chooses a new goal pose explicitly when using the \textit{waypoint navigation} method or implicitly in the \textit{early stopping} method. If the vehicle's distance to the goal $s\textsubscript{g}$ is smaller than a certain threshold $s\textsubscript{lim}$, the path is planned to the final goal pose and no navigation method is used.

\subsection{Replanning Method}
\begin{figure}[tbp]
\vspace{2mm}
	\centering
	\def\svgwidth{0.75\linewidth}
	\graphicspath{{img}}
\begingroup%
  \makeatletter%
  \providecommand\color[2][]{%
    \errmessage{(Inkscape) Color is used for the text in Inkscape, but the package 'color.sty' is not loaded}%
    \renewcommand\color[2][]{}%
  }%
  \providecommand\transparent[1]{%
    \errmessage{(Inkscape) Transparency is used (non-zero) for the text in Inkscape, but the package 'transparent.sty' is not loaded}%
    \renewcommand\transparent[1]{}%
  }%
  \providecommand\rotatebox[2]{#2}%
  \newcommand*\fsize{\dimexpr\f@size pt\relax}%
  \newcommand*\lineheight[1]{\fontsize{\fsize}{#1\fsize}\selectfont}%
  \ifx\svgwidth\undefined%
    \setlength{\unitlength}{203.24408584bp}%
    \ifx\svgscale\undefined%
      \relax%
    \else%
      \setlength{\unitlength}{\unitlength * \real{\svgscale}}%
    \fi%
  \else%
    \setlength{\unitlength}{\svgwidth}%
  \fi%
  \global\let\svgwidth\undefined%
  \global\let\svgscale\undefined%
  \makeatother%
  \begin{picture}(1,0.35912495)%
    \lineheight{1}%
    \setlength\tabcolsep{0pt}%
    \put(0,0){\includegraphics[width=\unitlength,page=1]{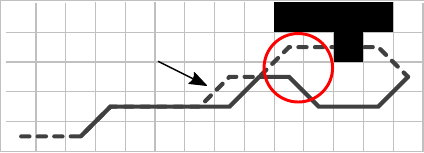}}%
    \put(0.37208424,0.23316461){\makebox(0,0)[t]{\lineheight{1.25}\smash{\begin{tabular}[t]{c}$d<d\textsubscript{div}$\end{tabular}}}}%
    \put(0,0){\includegraphics[width=\unitlength,page=2]{div_paths.pdf}}%
    \put(0.64785813,0.01673198){\makebox(0,0)[t]{\lineheight{1.25}\smash{\begin{tabular}[t]{c}$d>d\textsubscript{div}$\end{tabular}}}}%
    \put(0,0){\includegraphics[width=\unitlength,page=3]{div_paths.pdf}}%
  \end{picture}%
\endgroup%
\\
 \vspace{1mm}
\centering
\begin{minipage}{0.8\columnwidth}
\raggedleft
\begin{tikzpicture}
\begin{axis}[axis line style={draw=none}, tick style={draw=none}, yticklabels={,,}, xticklabels={,,}, width=3cm, height=1.7cm, legend columns=3]
\addplot[color=red, forget plot]{exp(x)};
\pgfplotsset{my area legend/.style={
  legend image code/.code={\fill[#1] (0cm,-0.2cm) rectangle (0.35cm,0.2cm);}
  }
}
\addlegendimage{dashed, gray}
\addlegendentry{prev. A*-path}
\addlegendimage{gray}
\addlegendentry{A*-path}
\addlegendimage{my area legend, black}
\addlegendentry{obstacles}
\end{axis}
\end{tikzpicture}
\end{minipage} 
	\caption{Detection of diverging A*-paths. The first deviation of the two paths is not significant and is therefore ignored. The second deviation, marked by the red circle, is significant.}
	\label{fig:div_paths}
\end{figure}

In this section, we introduce a replanning scheme that can handle unexplored environments efficiently and improves the stability of the overall system on replanning. This is achieved by detecting significant environment changes. The method is visualized in Fig.~\ref{fig:div_paths}. 
As mentioned in Sec.~\ref{subsec:nav}, a two-dimensional A*-path is extracted from the already calculated distance heuristic. Now, the positions of two A*-paths from subsequent timesteps are compared. At first, the paths are matched followed by the calculation of the element-wise distance $d$ of the A*-paths. If $d > d\textsubscript{div}$, the paths are called diverging and hence the environment has changed significantly, causing the two-dimensional path to find a different route to the goal position. This means that the previously generated path of the actual path planner should be replanned prior to the point of divergence to account for the changed environment. The distance along the path to this point of divergence is called $s\textsubscript{div}$ in the following.

In addition to that, the path should also be replanned if it collides with previously undetected obstacles. The distance to a collision is denoted by $s\textsubscript{coll}$. If no collision or divergence is present, the majority of the path can be reused. Hence, the distance to replan from is only bounded by the path length $s\textsubscript{path}$.
All distances stated above are calculated along the continuous or the two-dimensional A*-path, respectively. Further, an additional factor $\alpha$ is introduced that downscales the distance $s\textsubscript{plan}$ to replan. This is necessary as it was observed that especially the last part of a path is not necessarily a part of the optimal path to the final goal.
The pose from which the path should be replanned therefore lies at the distance
\begin{equation}
	s\textsubscript{plan} = \alpha \cdot \min\{ s\textsubscript{path}, s\textsubscript{coll}, s\textsubscript{div}\} \quad \text{with} \quad0 < \alpha < 1.
\end{equation}
To conclude, if the two-dimensional A*-paths diverge between two timesteps, a replanning is triggered. The start pose from where the Hybrid A* algorithm replans is the pose that lies at a distance of $s\textsubscript{plan}$ on the previously calculated continuous path.  
This replanning method together with the navigation method is called guided in the following section.
\subsection{Switching between System Models} \label{subsec:ra}
\begin{figure}[tbp]
\vspace{2mm}
 	\graphicspath{{img}}
	\centering
    \begin{minipage}{0.49\columnwidth}
    \def\svgwidth{1.0\columnwidth}
\begingroup%
  \makeatletter%
  \providecommand\color[2][]{%
    \errmessage{(Inkscape) Color is used for the text in Inkscape, but the package 'color.sty' is not loaded}%
    \renewcommand\color[2][]{}%
  }%
  \providecommand\transparent[1]{%
    \errmessage{(Inkscape) Transparency is used (non-zero) for the text in Inkscape, but the package 'transparent.sty' is not loaded}%
    \renewcommand\transparent[1]{}%
  }%
  \providecommand\rotatebox[2]{#2}%
  \newcommand*\fsize{\dimexpr\f@size pt\relax}%
  \newcommand*\lineheight[1]{\fontsize{\fsize}{#1\fsize}\selectfont}%
  \ifx\svgwidth\undefined%
    \setlength{\unitlength}{237.75506207bp}%
    \ifx\svgscale\undefined%
      \relax%
    \else%
      \setlength{\unitlength}{\unitlength * \real{\svgscale}}%
    \fi%
  \else%
    \setlength{\unitlength}{\svgwidth}%
  \fi%
  \global\let\svgwidth\undefined%
  \global\let\svgscale\undefined%
  \makeatother%
  \begin{picture}(1,0.90706732)%
    \lineheight{1}%
    \setlength\tabcolsep{0pt}%
    \put(0,0){\includegraphics[width=\unitlength,page=1]{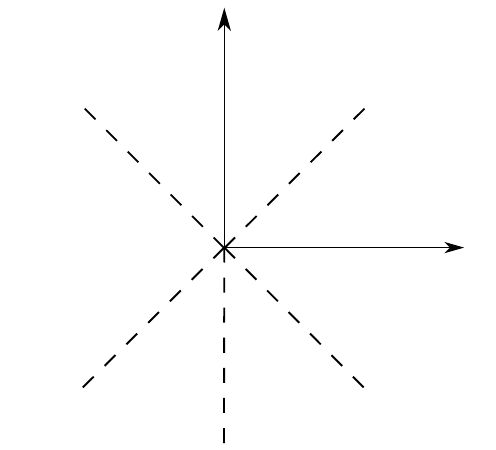}}%
    \put(0.87515993,0.6858914){\color[rgb]{0,0,0}\makebox(0,0)[lt]{\lineheight{1.25}\smash{\begin{tabular}[t]{l}$n \cdot \Delta\Phi$\\\end{tabular}}}}%
    \put(0,0){\includegraphics[width=\unitlength,page=2]{motion_primitives.pdf}}%
    \put(0.91340168,0.3457535){\makebox(0,0)[lt]{\lineheight{1.25}\smash{\begin{tabular}[t]{l}$x$\end{tabular}}}}%
    \put(0.37118808,0.86726587){\makebox(0,0)[lt]{\lineheight{1.25}\smash{\begin{tabular}[t]{l}$y$\\\end{tabular}}}}%
    \put(0.87515993,0.6858914){\color[rgb]{0,0,0}\makebox(0,0)[lt]{\lineheight{1.25}\smash{\begin{tabular}[t]{l}$n \cdot \Delta\Phi$\\\end{tabular}}}}%
    \put(0,0){\includegraphics[width=\unitlength,page=3]{motion_primitives.pdf}}%
  \end{picture}%
\endgroup%

    \end{minipage}
    \begin{minipage}{0.49\columnwidth}
    \def\svgwidth{1.0\linewidth}
\begingroup%
  \makeatletter%
  \providecommand\color[2][]{%
    \errmessage{(Inkscape) Color is used for the text in Inkscape, but the package 'color.sty' is not loaded}%
    \renewcommand\color[2][]{}%
  }%
  \providecommand\transparent[1]{%
    \errmessage{(Inkscape) Transparency is used (non-zero) for the text in Inkscape, but the package 'transparent.sty' is not loaded}%
    \renewcommand\transparent[1]{}%
  }%
  \providecommand\rotatebox[2]{#2}%
  \newcommand*\fsize{\dimexpr\f@size pt\relax}%
  \newcommand*\lineheight[1]{\fontsize{\fsize}{#1\fsize}\selectfont}%
  \ifx\svgwidth\undefined%
    \setlength{\unitlength}{368.50393701bp}%
    \ifx\svgscale\undefined%
      \relax%
    \else%
      \setlength{\unitlength}{\unitlength * \real{\svgscale}}%
    \fi%
  \else%
    \setlength{\unitlength}{\svgwidth}%
  \fi%
  \global\let\svgwidth\undefined%
  \global\let\svgscale\undefined%
  \makeatother%
  \begin{picture}(1,0.69230769)%
    \lineheight{1}%
    \setlength\tabcolsep{0pt}%
    \put(0,0){\includegraphics[width=\unitlength,page=1]{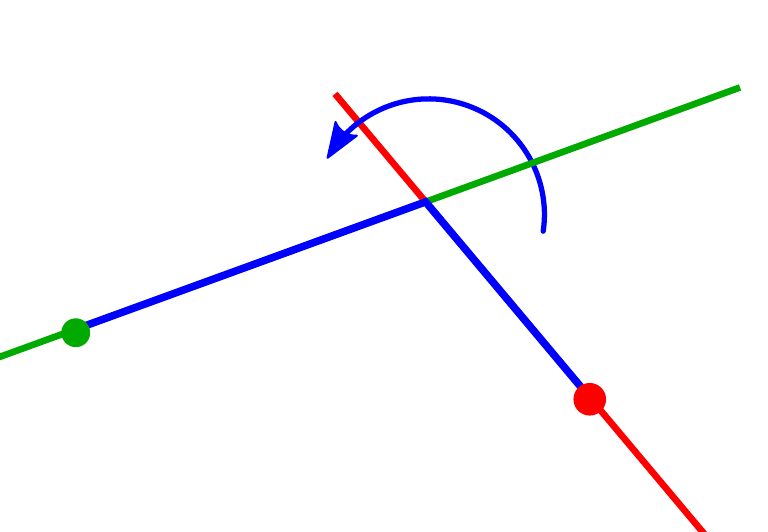}}%
    \put(0.10169781,0.33631058){\color[rgb]{0,0,0}\makebox(0,0)[lt]{\lineheight{1.25}\smash{\begin{tabular}[t]{l}$P\textsubscript{c}$\end{tabular}}}}%
    \put(0.59910424,0.59248422){\color[rgb]{0,0,0}\makebox(0,0)[lt]{\lineheight{1.25}\smash{\begin{tabular}[t]{l}$\Delta\Phi$\end{tabular}}}}%
    \put(0.80435416,0.22440108){\color[rgb]{0,0,0}\makebox(0,0)[lt]{\lineheight{1.25}\smash{\begin{tabular}[t]{l}$P\textsubscript{g}$\end{tabular}}}}%
  \end{picture}%
\endgroup%

    \end{minipage}\\
    \begin{minipage}{0.49\columnwidth}
        \centering
    \footnotesize{a)}
    \end{minipage}
    \begin{minipage}{0.49\columnwidth}
    \centering
    \footnotesize{b)}
    \end{minipage}
	\caption{a): The extended motion primitives are shown in blue along with the standard ones in red. b): Geometric extension: The intersection of the line segment of the current node's pose $P\textsubscript{c}$ (green) with the line segment of the goal's pose $P\textsubscript{g}$ (red) defines the point of rotation. The path to drive is shown in blue.}
    \label{fig:add_motion_primitives}
\end{figure}

In this section, the guided Hybrid A* algorithm is further modified to handle the extended maneuverability of the U-Shift concept vehicle shown in Fig.~\ref{fig:system} c).
Therefore, the two major planning concepts of the algorithm are modified in the following ways: 1) Additional motion primitives are inserted during the exploration steps of the graph. 2) Similar to the Reeds-Shepp (RS) curves \cite{rs} of the original approach, a geometric extension is introduced to reach the final pose by rotating on the rear axis.
Fig.~\ref{fig:add_motion_primitives} visualizes both modifications.

During the graph exploration, the additional motion primitives are applied with a frequency $f\textsubscript{ext}$. The angle of rotation is discretized in steps of $\Delta\Phi$. Hence, only multiples $n$ of $\Delta\Phi$ are used. For every rotation around the rear axis, the vehicle must switch to its second model. Hence, each rotation is penalized by an additional term of movement costs to account for the time it takes to execute the model change. It must turn the wheels to the inside, rotate and then turn the wheels back straight which takes $\approx\qty{5}{s}$ not counting the time for the rotation itself which depends on the angle.

In addition to this, a geometric extension is used. Here, two line segments with their center on top of the pose of the current node and the goal pose are generated. If they intersect, the point of intersection specifies the point to turn on the rear axis. The angular difference $\Delta\Phi$ specifies the angle to rotate. During the execution of the Hybrid A* algorithm, this extension is applied whenever the RS extension is executed. This method is the equivalent of executing the RS extension with infinite curvature. 
\section{EVALUATION} \label{sec:eval}
The proposed methods are now evaluated. First, some evaluation details are mentioned. Second, the novel navigation approach is compared to the conventional navigation scheme. After this, the complete guiding method as well as the extended version of the Hybrid A* planner using the additional motion primitives is evaluated.
\subsection{Evaluation Details}
For this paper, the standard Hybrid A* method was implemented and adapted with the proposed methods. It is combined with the sparse collision checking method developed by \cite{coll_check}. The coordinates of a varying number of disks are precalculated, whose number and positions are based on the existence and size of the transported capsule. The paths of the standard Hybrid A* algorithm are always generated to the final goal and are only replanned if the path collides within a certain distance $s\textsubscript{coll}$. This was used to allow the algorithm to explore the environment slightly longer before it replans. All experiments were done on an Intel(R) Core(TM) i9-7900X CPU @ 3.30GHz.

The proposed methods are deployed on a research vehicle of Ulm University and tested on a parking lot for the generation of qualitative results. The vehicle is equipped with several LIDAR sensors and a highly precise Inertial Measurement Unit (IMU) using DGNSS (Differential Global Navigation Satellite System) which enables the precise mapping of its environment which is based on the approach proposed in \cite{adaptive_patched_grid_mapping}.
The path is passed to an optimization-based longitudinal trajectory planner proposed in \cite{traj_planning_jona}. Table~\ref{tab:params} shows the parameters used in the proposed guidance and extension method together with the main parameters of the Hybrid A* algorithm.

Furthermore, the proposed algorithm is investigated in simulations to generate reproducible results. The simulation framework emulates the exploration of a previously unknown environment by applying raytracing on a predefined obstacle map followed by a mapping step.
\begin{table}[tbp]
\vspace{2mm}
	\caption{Parameters}
	\label{tab:params}
	\begin{center}
		\begin{tabular}{c | l | c}
			\toprule
			Parameter & Description & Value \\
			\midrule
			$s\textsubscript{w}$ & dist. of \textit{early stopping} & \qty{55}{\meter} \\
			$s\textsubscript{lim}$ & dist. over which \textit{early stopping} is applied & \qty{60}{\meter} \\
        $d\textsubscript{div}$ & allowed dist. of divergence & \qty{5}{\meter} \\   
			$\alpha$ & factor to scale dist. to new start pose & 0.5 \\
			$s\textsubscript{coll}$ & allowed dist. to collision & \qty{20}{\meter} \\
            \midrule
            $r\textsubscript{xy}$ & spatial resolution of the Hybrid A* & \qty{0.625}{\meter} \\
            $r\textsubscript{$\Phi$}$ & yaw resolution of the Hybrid A* & \qty{10}{\degree} \\
            $\delta\textsubscript{max}$ & max steering angle & \qty{31.51}{\degree} \\
            $r\textsubscript{grid}$ & spatial resolution of the grid map & \qty{0.15625}{\meter} \\
			\bottomrule
		\end{tabular}
	\end{center}
   \vspace{-3mm}
\end{table}%
$s\textsubscript{w}$ and $s\textsubscript{lim}$ control if and when the \textit{early stopping} is applied. The smaller the distance, the less processing power is needed. However, with decreasing values, the planning algorithm plans more locally which can be a problem in situations that need complex maneuvers. $d\textsubscript{div}$ tunes the sensitivity to environmental changes. Here, this value was chosen so that no divergent A*-paths occur in known environments. With rising $\alpha$, less processing power is needed but large environment changes might require a replanning from further back to find an optimal path.
\subsection{Guidance Method}\label{sec:guidance}
The next section compares the novel navigation method from Sec.~\ref{subsec:nav} with the baseline and demonstrates one pitfall that can arise with already published navigation approaches. Then the impact of the combined usage of the navigation and replanning method is evaluated.
\subsubsection{Comparison of Navigation Methods}
\begin{figure}[tbp]
\vspace{2mm}
       \begin{minipage}{0.15\columnwidth}
      \centering
      \footnotesize{a)}
        \end{minipage}
      \begin{minipage}{0.7\columnwidth}
          \frame{\includegraphics[trim={9cm 4.5cm 1.5cm 4.5cm},clip, width=1.0\columnwidth]{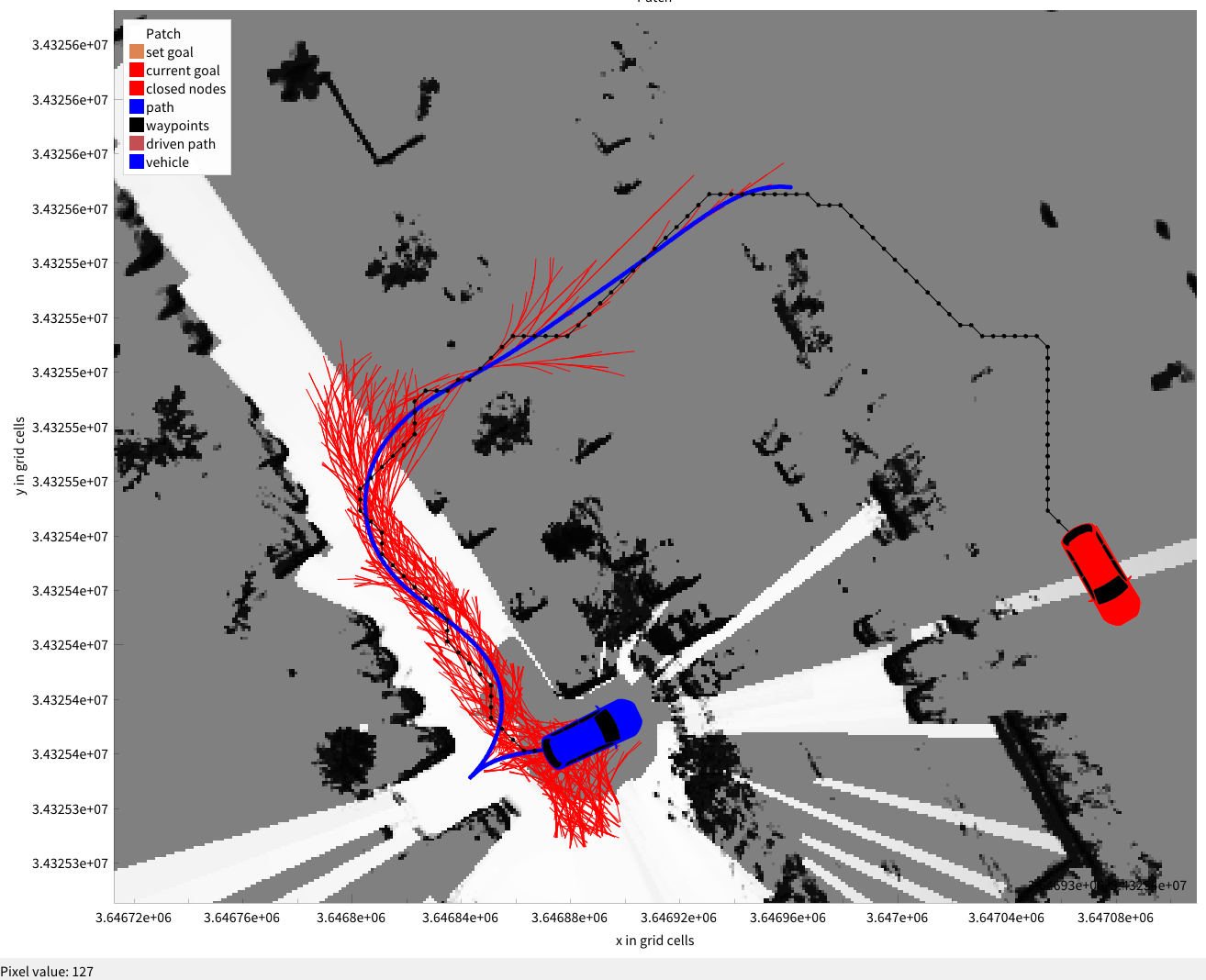}}%
     \end{minipage}\\
    \vspace{1mm}\\
      \begin{minipage}{0.15\columnwidth}
      \centering
      \footnotesize{b)}
     \end{minipage}
    \begin{minipage}{0.7\columnwidth}
      \frame{\includegraphics[trim={10cm 7cm 1.5cm 3cm},clip, width=1.0\columnwidth]{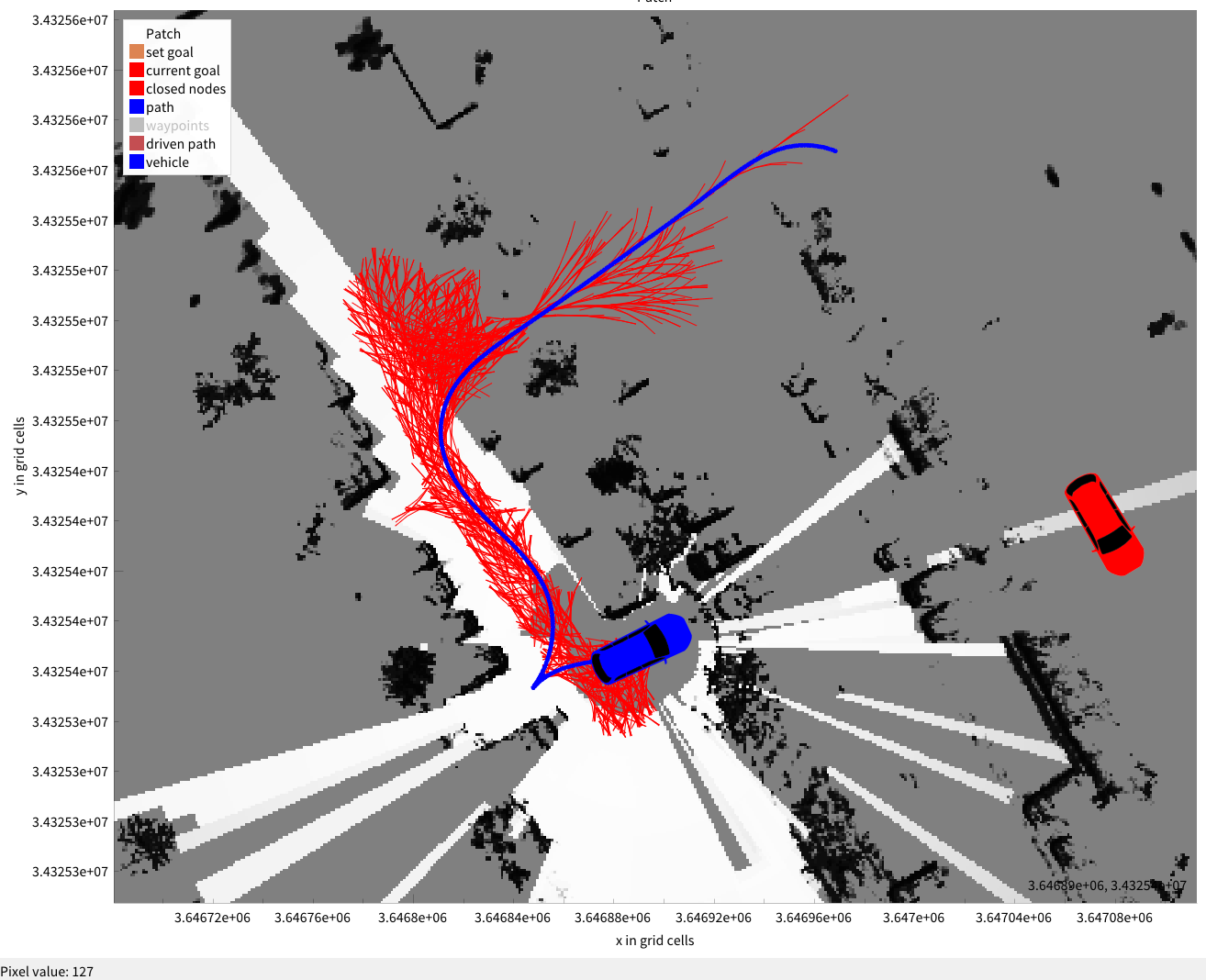}}%
     \end{minipage}\\
\centering
\begin{minipage}{0.86\columnwidth}
\raggedleft
\begin{tikzpicture}
\begin{axis}[axis line style={draw=none}, tick style={draw=none}, yticklabels={,,}, xticklabels={,,}, width=3cm, height=1.7cm, legend columns=3]
\addplot[color=red, forget plot]{exp(x)};
\addlegendimage{no markers, blue}
\addlegendentry{path}
\addlegendimage{no markers, red}
\addlegendentry{explored graph}
\addlegendimage{no markers, dashed, black}
\addlegendentry{A*-path}
\end{axis}
\end{tikzpicture}
\end{minipage}
	\caption{Planned paths of the research vehicle of Ulm University (blue) to the goal pose (red) with the explored graph. a): Guidance by \textit{waypoint navigation}. b): Guidance by \textit{early stopping}.}
	\label{fig:path_to_waypoint}
\end{figure}
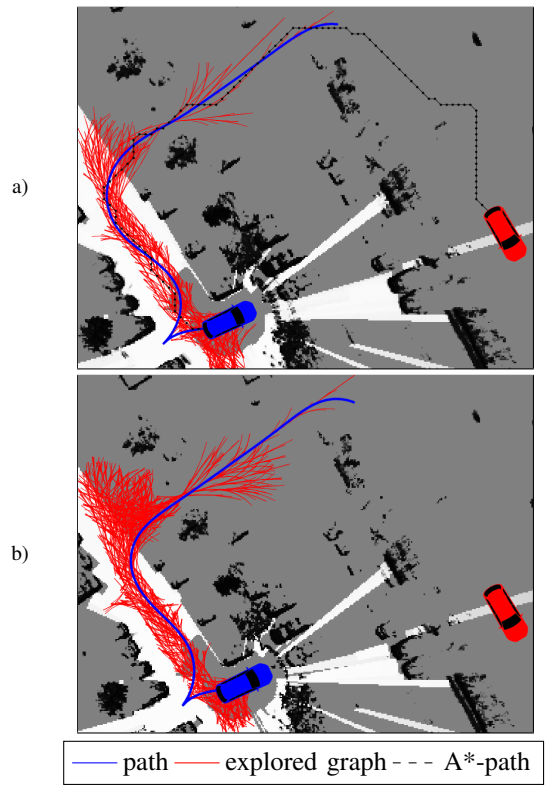

\begin{figure}[tbp]
\vspace{2mm}
       \begin{minipage}{0.15\columnwidth}
      \centering
      \footnotesize{a)}
     \end{minipage}
     \begin{minipage}{0.65\columnwidth}
      \frame{\includegraphics[trim={11cm 10.5cm 8.5cm 9.2cm},clip, width=1.0\columnwidth]{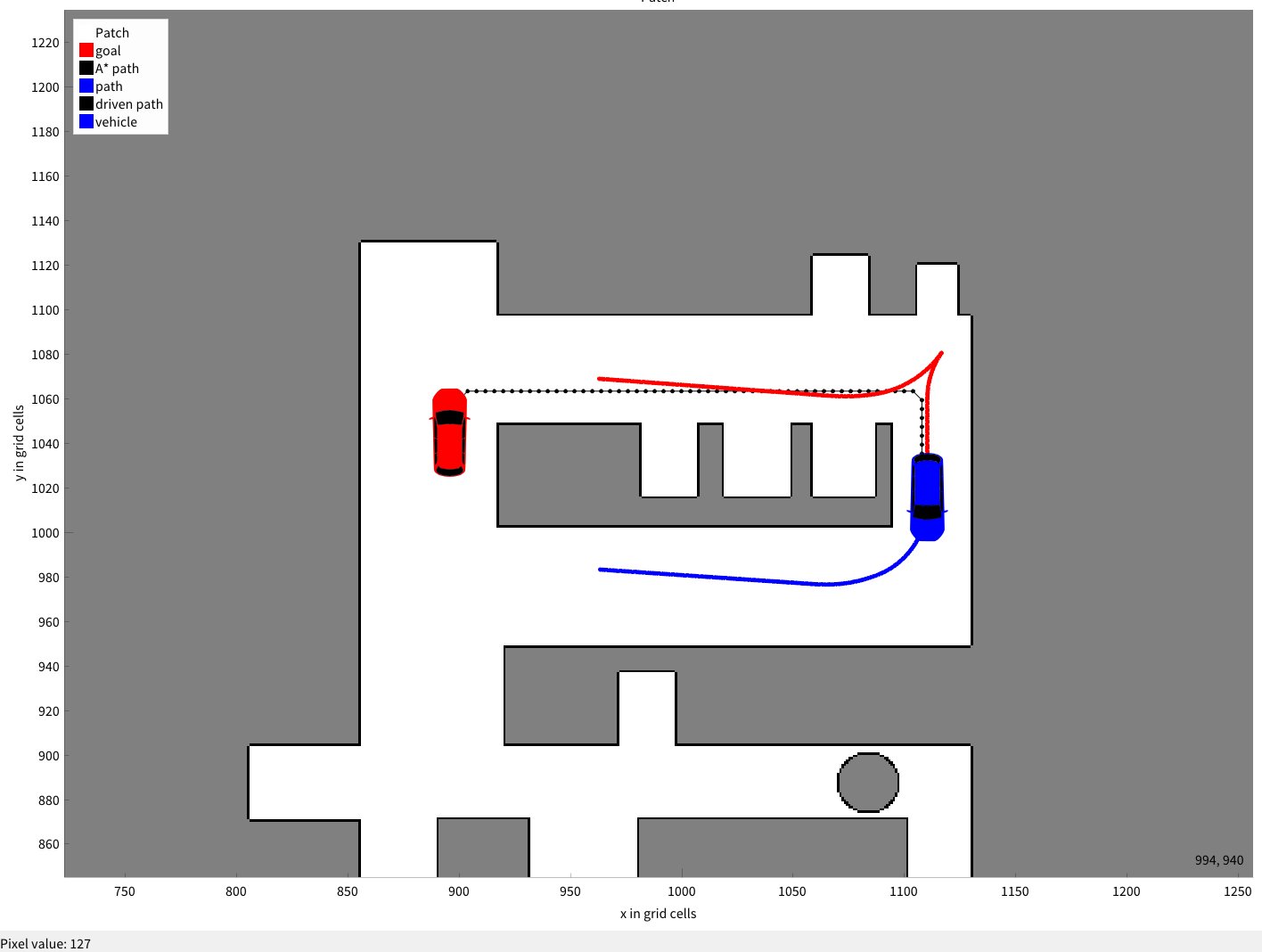}}%
     \end{minipage}\\
    \vspace{1mm}\\
      \begin{minipage}{0.15\columnwidth}
      \centering
      \footnotesize{b)}
     \end{minipage}
    \begin{minipage}{0.65\columnwidth}
     \frame{\includegraphics[trim={5cm 10.5cm 3cm 11.1cm},clip, width=1.0\columnwidth]{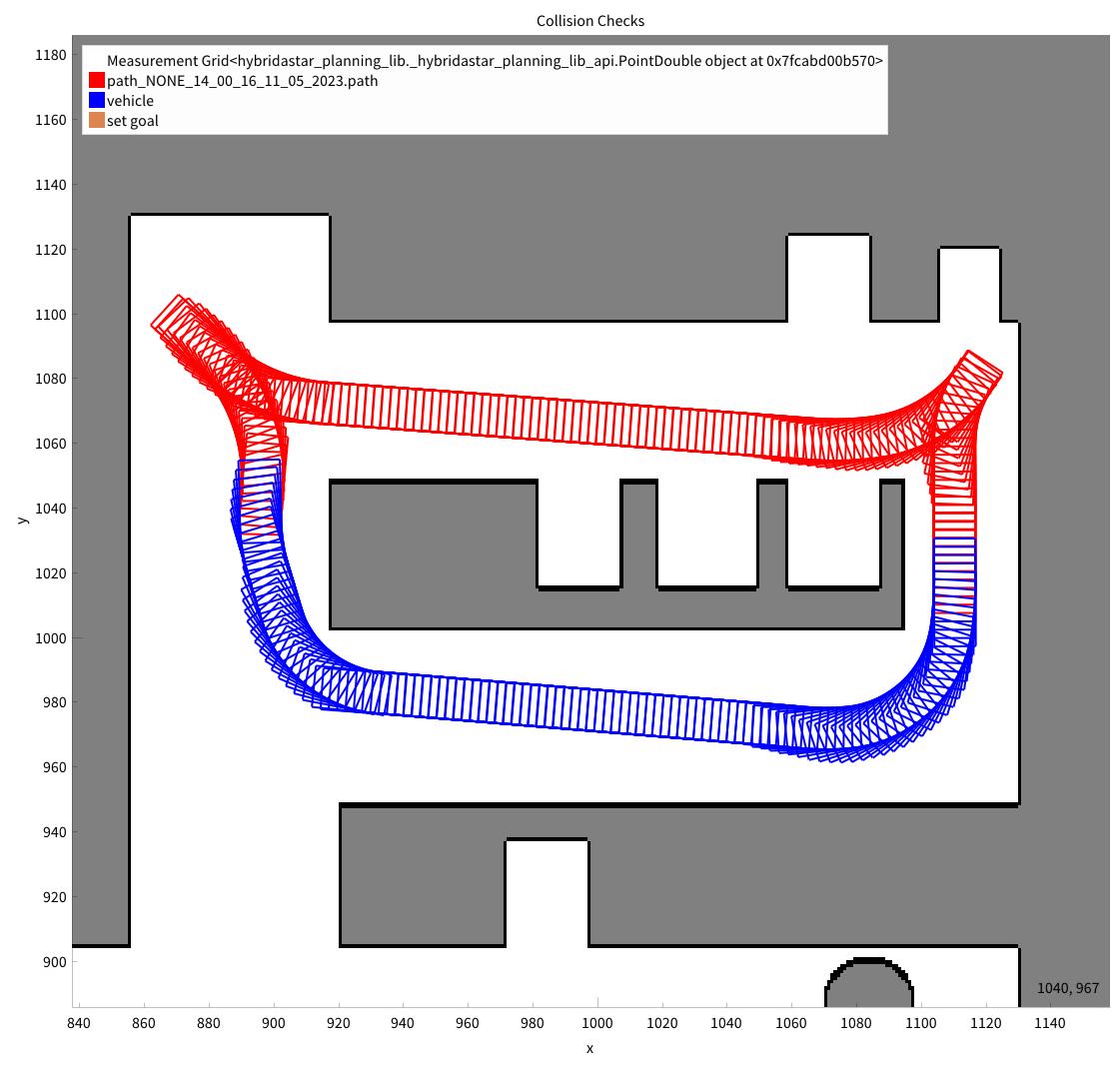}}%
     \end{minipage}\\
 \raggedright
\begin{minipage}{0.91\columnwidth}
\raggedleft
\begin{tikzpicture}
\begin{axis}[axis line style={draw=none}, tick style={draw=none}, yticklabels={,,}, xticklabels={,,}, width=3cm, height=1.7cm, legend columns=2]
\addplot[color=red, forget plot]{exp(x)};
\addlegendimage{no markers, red}
\addlegendentry{waypoint navigation}
\addlegendimage{no markers, blue}
\addlegendentry{early stopping}
\end{axis}
\end{tikzpicture}
\end{minipage}
	\caption{The ego pose in blue and the goal pose in red. a) Planned paths of the \textit{waypoint navigation} and the \textit{early stopping} method.
    b): Resulting paths. The \textit{early stopping} method does not create unnecessary movements.}
	\label{fig:nav_comparison}
\end{figure}
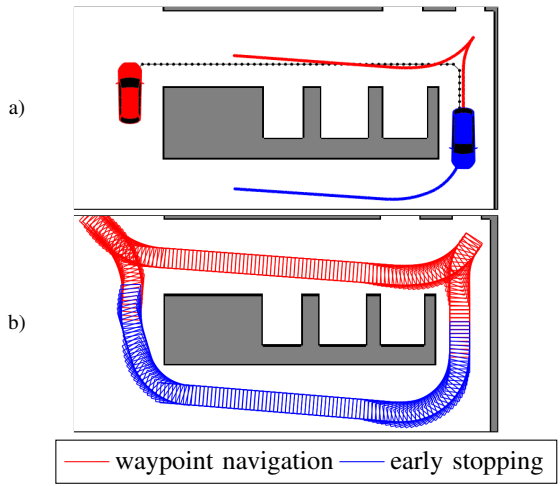

Fig.~\ref{fig:path_to_waypoint} shows the qualitative results of the path planner using the \textit{waypoint navigation} method and the \textit{early stopping} method on real-world data on a parking lot. Here, the resulting paths of the two methods are nearly identical.
However, it was observed that conventional navigation methods like the introduced \textit{waypoint navigation} method or the approaches presented in \cite{way1, way2, twostage_hybrid} are inherently prone to the following problem. It can occur that the navigation approach takes a different direction around an obstacle than the detailed path planning algorithm would. This is due to the higher dimension of the detailed planner and
additional cost terms e.g. for steering.
This is visualized in Fig.~\ref{fig:nav_comparison}. Here, the \textit{waypoint navigation} method generates a sub-goal that does not lie near the optimal path considering the vehicle's orientation. This leads to the sub-optimal resulting path of the \textit{waypoint navigation} method in red with two direction changes. The \textit{early stopping} method is superior to methods of this kind as it does not have these problems inherently. It does not aim for an explicit goal but stops when the desired state, the reduction of the distance heuristic, is reached. Hence, it generates a different path as it also considers the vehicle's orientation. The \textit{early stopping} method results in an improved path without direction changes.

\subsubsection{Simulation Results}
The goal of this section is to investigate if the guided version of the Hybrid A* algorithm has a decreased runtime as expected and to what extent the driven path still fulfills the requirements of smoothness, clearance, and length of the path.

\begin{figure}[tbp]
\vspace{2mm}
      \begin{minipage}{0.15\columnwidth}
      \centering
      \footnotesize{a)}
     \end{minipage}
     \begin{minipage}{0.7\columnwidth}
     \frame{\includegraphics[angle=270, trim={17.5cm 2cm 24.5cm 1cm},clip, width=1.0\columnwidth]{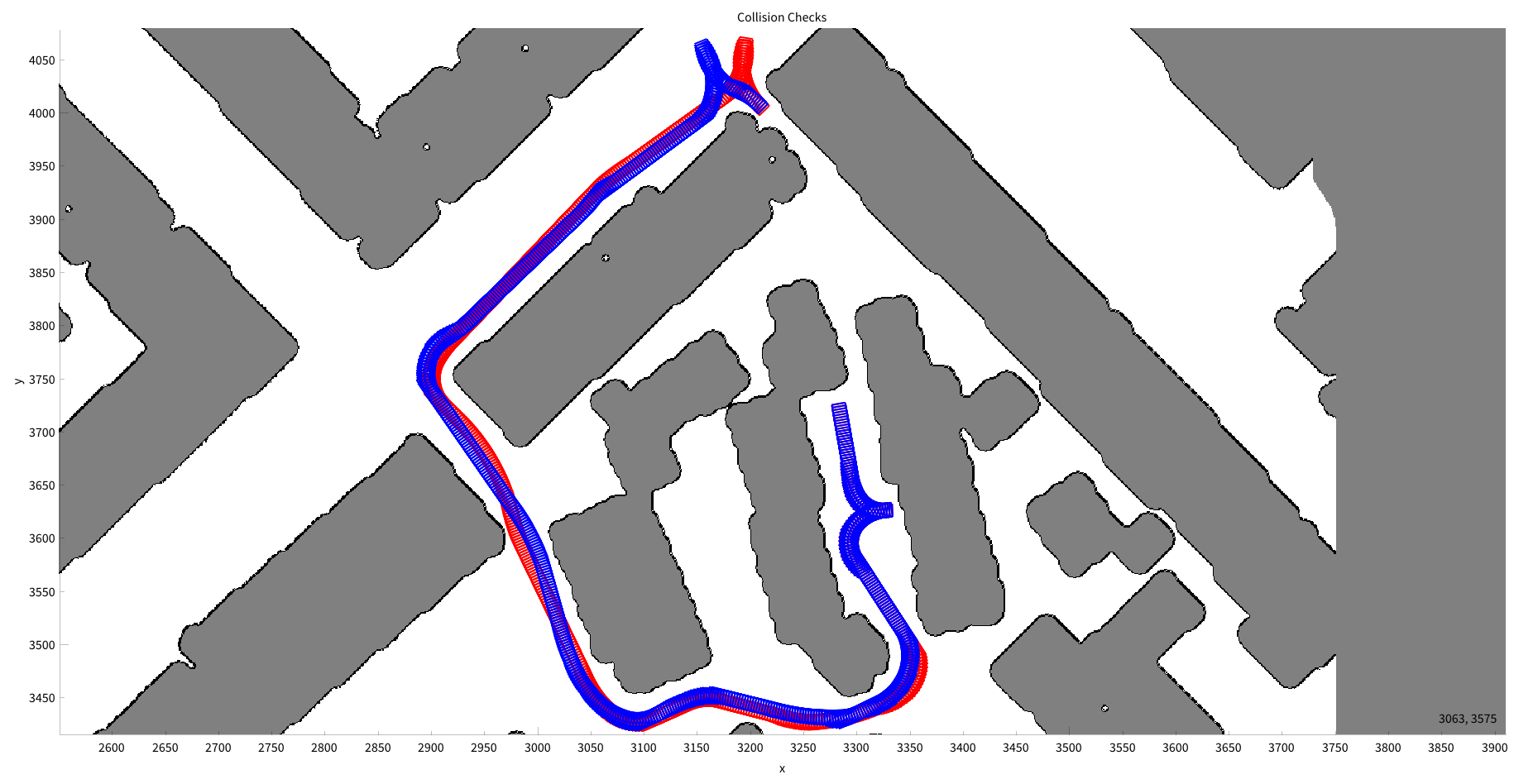}}%
     \end{minipage}\\
    \vspace{1mm}\\
      \begin{minipage}{0.15\columnwidth}
      \centering
      \footnotesize{b)}
     \end{minipage}
    \begin{minipage}{0.7\columnwidth}
     \frame{\includegraphics[angle=270, trim={16.5cm 2cm 25.5cm 1cm},clip, width=1.0\columnwidth]{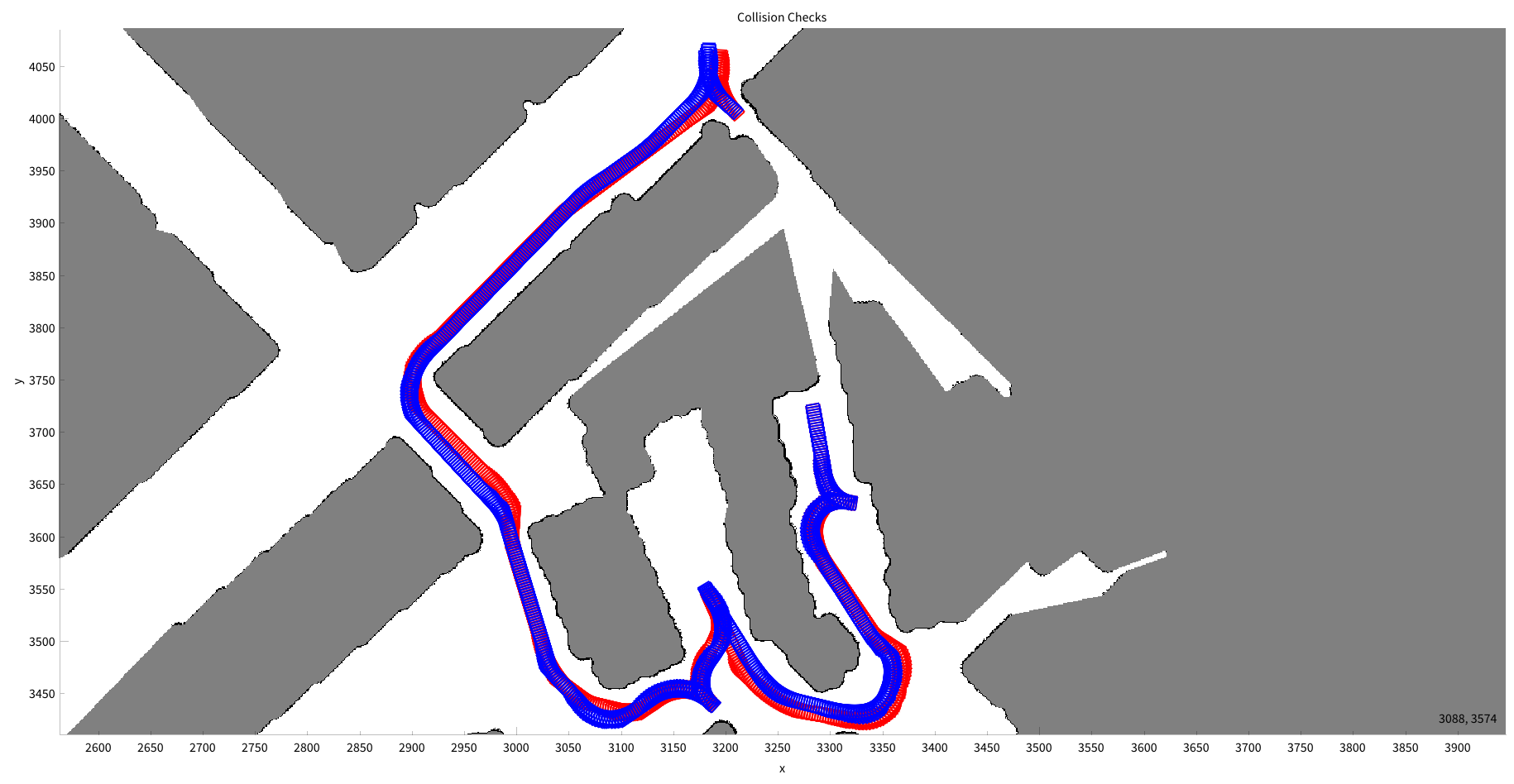}}%
     \end{minipage}\\
 \raggedright
\begin{minipage}{0.84\columnwidth}
\raggedleft
\begin{tikzpicture}
\begin{axis}[axis line style={draw=none}, tick style={draw=none}, yticklabels={,,}, xticklabels={,,}, width=3cm, height=1.7cm, legend columns=3]
\addplot[color=red, forget plot]{exp(x)};
\addlegendimage{area legend, red}
\addlegendentry{standard HA*}
\addlegendimage{area legend, blue}
\addlegendentry{guided HA*}
\end{axis}
\end{tikzpicture}
\end{minipage}
	\caption{Qualitative comparison of the generated paths. The paths are very similar in previously known a) and unexplored environments b).}
	\label{fig:path_comparison}
\end{figure}
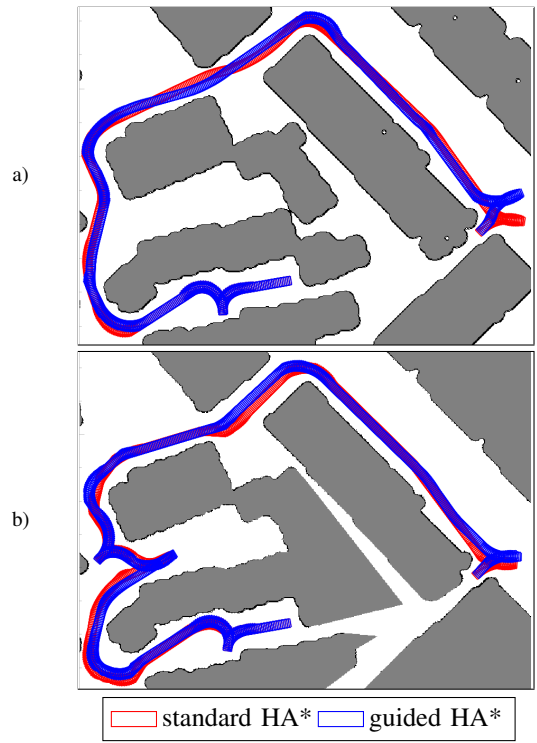

Here, both methods are compared qualitatively in two large environments shown in Fig.~\ref{fig:path_comparison}. One is previously known, and the other one is unexplored.  
In both environments, the paths of the standard and guided methods differ only insignificantly. The generation of the two figures can be observed in the following two videos
\footnote{\href{https://youtu.be/k8ezypm78WQ}{https://youtu.be/k8ezypm78WQ} and \href{https://youtu.be/hHTfiry8gd0}{https://youtu.be/hHTfiry8gd0}}.
The runtimes are compared quantitatively in Table~\ref{tab:navigation_replanning_runtimes}. In the known environment, the guided approach executes the path planner multiple times compared to the standard approach which directly plans to the final goal. Furthermore, the guided method has a reduced maximum and average planning time. This was predictable, as the goal to plan to is always closer to the current position. However, the cumulative planning time is also reduced by a factor of $\approx1.6$. This shows the overall reduction of energy and processing power of the path planner. 

In the case of an unknown environment, the standard Hybrid A* algorithm must replan as well due to the discovered obstacles. Here, the runtimes behave similarly. The guided method has a lower runtime concerning the maximum, average, and cumulative runtimes. The cumulative runtime of the guided method is by a factor of $\approx2$ smaller than in the standard case. In addition to that, the average runtime is reduced by a factor of $\approx4.6$ which allows the path planner to react faster to environment changes and lowers the overall system latency.
\begin{table}[bp]
	\caption{Runtime comparison of standard and guided Hybrid A* in known and unknown environments}
	\label{tab:navigation_replanning_runtimes}
	\begin{center}
		\begin{tabular}{c  c | c | c  c  c | c }
			\toprule
			Env. & Method & numb. of & $t\textsubscript{max}$ & $t\textsubscript{cum}$ & $t\textsubscript{avg}$ & cum.  \\
            &  & exec.& \multicolumn{3}{c|}{in \si{\second}} & nodes\\
			\midrule
			\multirow{2}{*}{known} & std. & 1 & \multicolumn{3}{c|}{0.63} & 50049 \\
			& guided & 33 & 0.135 & \textbf{0.38} & 0.0115 & \textbf{40559} \\
			\midrule
			\multirow{2}{*}{unkn.} & std. & 44 & 0.38 & 3.8 & 0.087 & 413575\\
			& guided & 97 & \textbf{0.16} & \textbf{1.83} & \textbf{0.019} & \textbf{185872}\\
			\bottomrule
		\end{tabular}
	\end{center}
\end{table}%

Next, the generated paths are compared with respect to their smoothness and clearance to obstacles in Table~ \ref{tab:navigation_replanning_path}. One well-known metric for smoothness is the root mean square (RMS) of the change of curvature $\dot{\kappa}$ which was introduced in \cite{smoothness}. It is calculated by
\begin{equation}
	\dot{\kappa}\textsubscript{RMS} = \sqrt{\frac{1}{N}\sum_{i=0}^{N-1} \left(\frac{\kappa_{i+1} - \kappa_{i}}{ds} \right)^{2}} = \sqrt{\frac{1}{N}\sum_{i=0}^{N-1} \dot{\kappa}^{2}}.
\end{equation}
Furthermore, the average and maximum proximity $p\textsubscript{avg}$ and $p\textsubscript{max}$ to obstacles is compared.
by extracting the maximum value of the Generalized Voronoi Diagram (GVD) \cite{hybridastar} of all corners of the vehicle. A GVD states the normed distance to obstacles with a value of 0 for a point far from or between obstacles and a value of 1 for positions within obstacles. At last, the length of the generated paths is compared. All metrics are calculated on the resulting driven path after the goal pose was reached. 
\begin{table}[btp]
\vspace{2mm}
	\caption{Path comparison of standard and guided Hybrid A* in \\known and unknown environments}
	\label{tab:navigation_replanning_path}
	\begin{center}
		\begin{tabular}{c  c | c  c  | c  c | c }
			\toprule
			Env. & Method & $\dot{\kappa}\textsubscript{RMS}$ & $\mid \dot{\kappa}\textsubscript{max} \mid$ & $p\textsubscript{max}$ & $p\textsubscript{avg}$ & $l$ \\
            & & \multicolumn{2}{c|}{in $1/\si{\meter}^2$} &\multicolumn{2}{c|}{-} & in \si{\meter} \\
			\midrule
			\multirow{2}{*}{known} & std.  & 0.39
			 & 9.64
			  & \textbf{0.68}
			   & \textbf{0.0027}
			    & 224
			     \\
			& guided & \textbf{0.32}
			 & 9.64
			  & 1.0
			   & 0.0044
			    & 224
			     \\
            \midrule
			\multirow{2}{*}{unknown} & std. & 0.53
			 & 11.28
			  & 1.0
			   & \textbf{0.0034}
			    & 255
			     \\
			& guided & \textbf{0.51}
			 & 11.28
			  & 1.0
			   & 0.0063
			    & 252
			     \\
			\bottomrule
		\end{tabular}
	\end{center}
\end{table}%
Table~\ref{tab:navigation_replanning_path} shows that paths generated by the standard Hybrid A* algorithm and the guided version are very similar in all metrics in known and unexplored environments. The guided method tends to create slightly smoother paths at the cost of a slightly worse clearance to obstacles. 

To summarize, the guidance method leads to shorter runtimes. It reduces the cumulative planning time up to a factor of $\approx2$ and reduces the average planning time up to a factor of $\approx4.6$. In known and previously unknown environments, the guided method leads to similar paths with respect to the smoothness, clearance, and length of the path. Hence, the proposed guidance method can be used to plan paths more efficiently.

\subsection{Extended System Model} \label{subsec:ra_eval}
This section compares the standard Hybrid A* algorithm against the extended version from Sec.~\ref{subsec:ra}.
\subsubsection{Parameter Tuning and Proof of Concept}
At first, the execution times are compared in Table~\ref{tab:extended_versus_vanilla} for a varying set of parameters to investigate if the algorithm is still real-time capable. The values are stated relative to the standard Hybrid A* algorithm.
\begin{table}[bp]
	\caption{Relative values of the extended against the standard Hybrid A* algorithm.}
	\label{tab:extended_versus_vanilla}
	\begin{center}
		\begin{tabular}{c | c c | c | c }
			\toprule
			Method & f\textsubscript{ext} & $\Delta\Phi$ & cum. & t\textsubscript{avg} \\
			 & every $x$ node & in \si{\degree} &  nodes & - \\
			\midrule
			standard & - & - & 1.0 & 1.0\\
			\midrule
			\multirow{4}{*}{extended} & 1 & \qty{45}{\degree} & 2.38
			 & 2.53 \\
			& 5 & \qty{45}{\degree} & 1.38
			 & 1.33 \\
			& 1 & \qty{90}{\degree} & 1.96
			 & 1.67 \\
			& \textbf{5} & $\mathbf{90}$\si{\degree} & \textbf{1.19}
			 & \textbf{1.14} \\
			\bottomrule
		\end{tabular}
	\end{center}
\end{table}%
With rising frequency $f\textsubscript{ext}$ of applying the extended motion primitives and finer $\Delta\Phi$, the execution time of the overall algorithm rises. This is caused by the increased amount of nodes that are explored. However, it can be seen that the execution times stay in the same order of magnitude. Furthermore, it can be observed that if the extended motion primitives are applied with $f\textsubscript{ext}=5$ and $\Delta\Phi=\qty{90}{\degree}$, the execution time increases only slightly compared to the standard Hybrid A* algorithm. Therefore, these parameters are used in the following experiments.

\begin{figure}[tbp]
\vspace{2mm}
        \begin{minipage}{0.1\columnwidth}
         \centering
        \footnotesize{a)}
        \end{minipage}
        \begin{minipage}{0.8\columnwidth}
        \includegraphics[trim={2.5cm 16.5cm 0 12cm},clip, width=1.0\columnwidth]{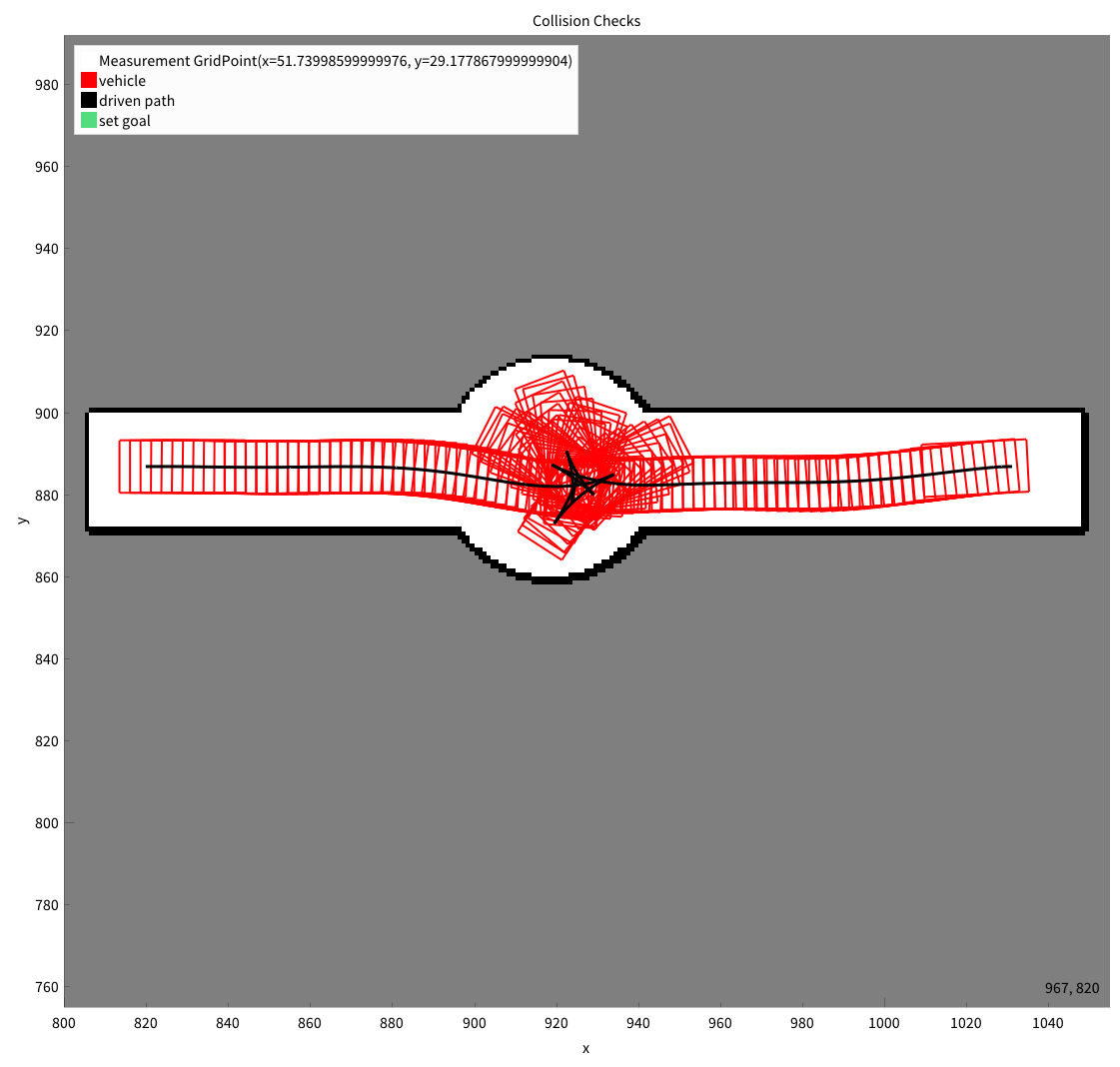}
        \end{minipage}\\
        \vspace{1mm}\\
        \begin{minipage}{0.1\columnwidth}
         \centering
        \footnotesize{b)}
        \end{minipage}
        \begin{minipage}{0.8\columnwidth}
        \includegraphics[trim={2.5cm 14.5cm 0 15.5cm},clip, width=1.0\columnwidth]{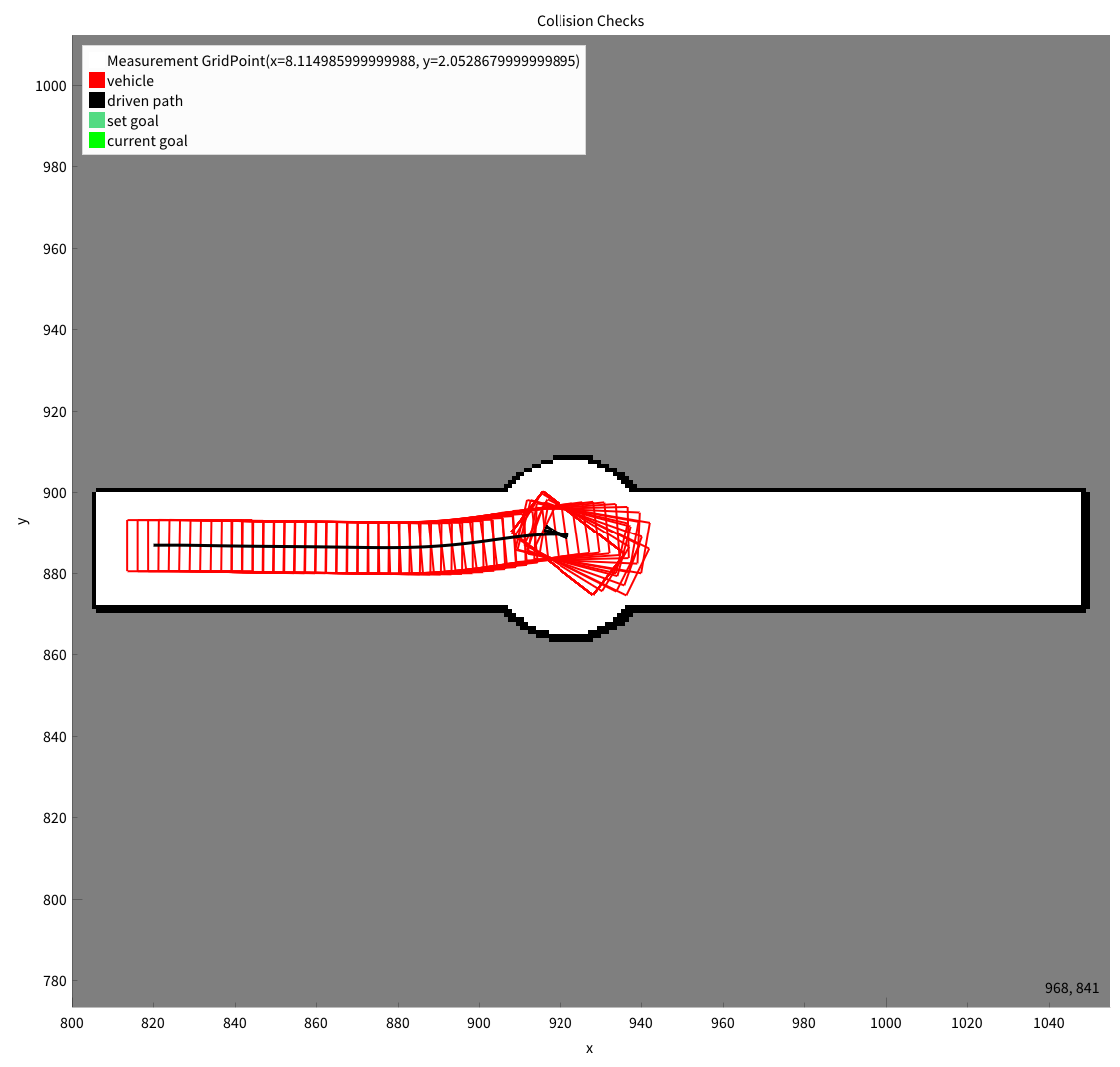}%
        \end{minipage}\\
        \vspace{1mm}\\
        \begin{minipage}{0.1\columnwidth}
         \centering
        \footnotesize{c)}
        \end{minipage}
        \begin{minipage}{0.8\columnwidth}
        \includegraphics[trim={2.5cm 14.4cm 0 14.7cm},clip, width=1.0\columnwidth]{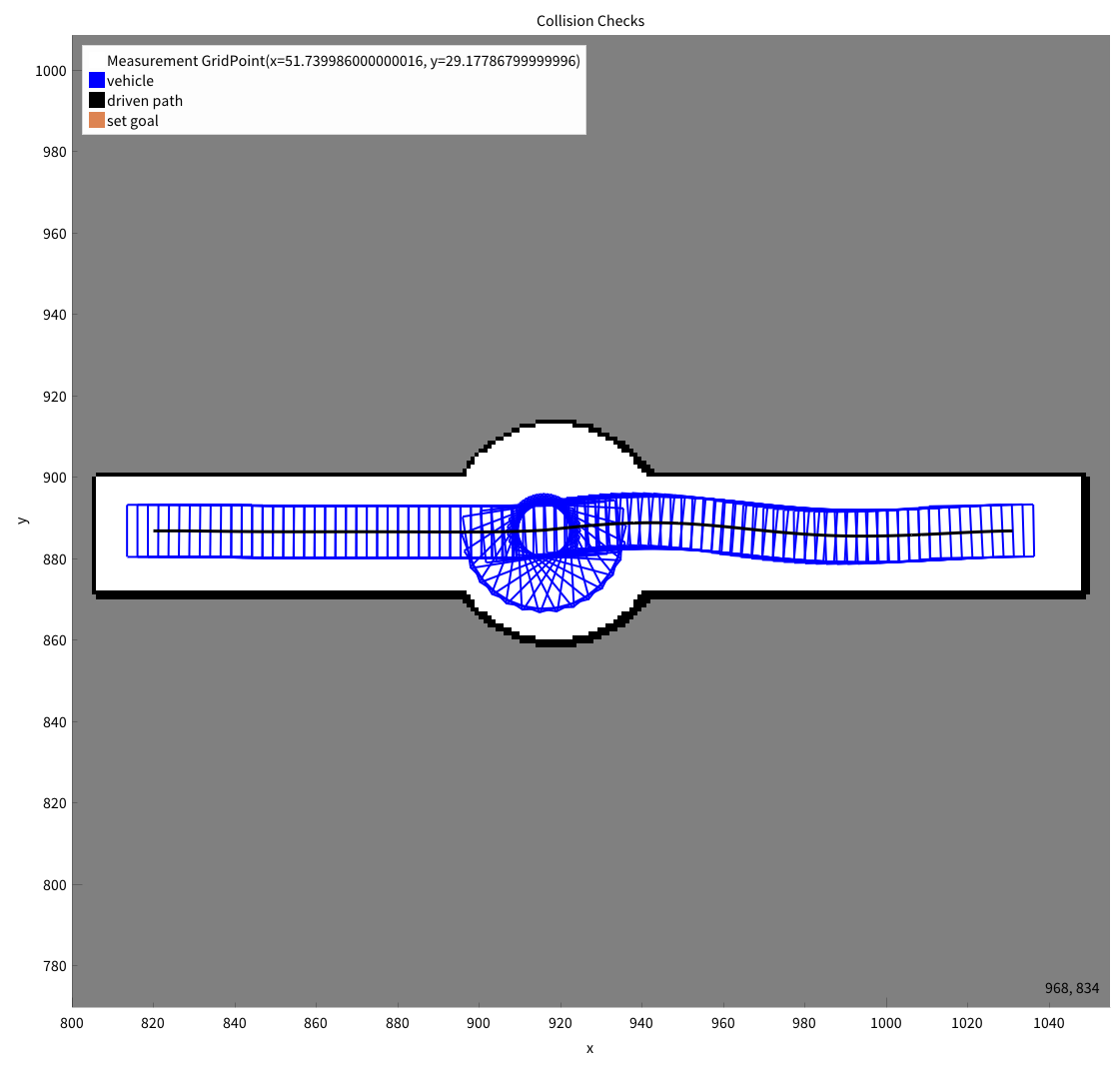}%
        \end{minipage}\\
        \vspace{1mm}\\
        \begin{minipage}{0.1\columnwidth}
         \centering
        \footnotesize{d)}
        \end{minipage}
        \begin{minipage}{0.8\columnwidth}
        \includegraphics[trim={2.5cm 17.2cm 0 13.2cm},clip, width=1.0\columnwidth]{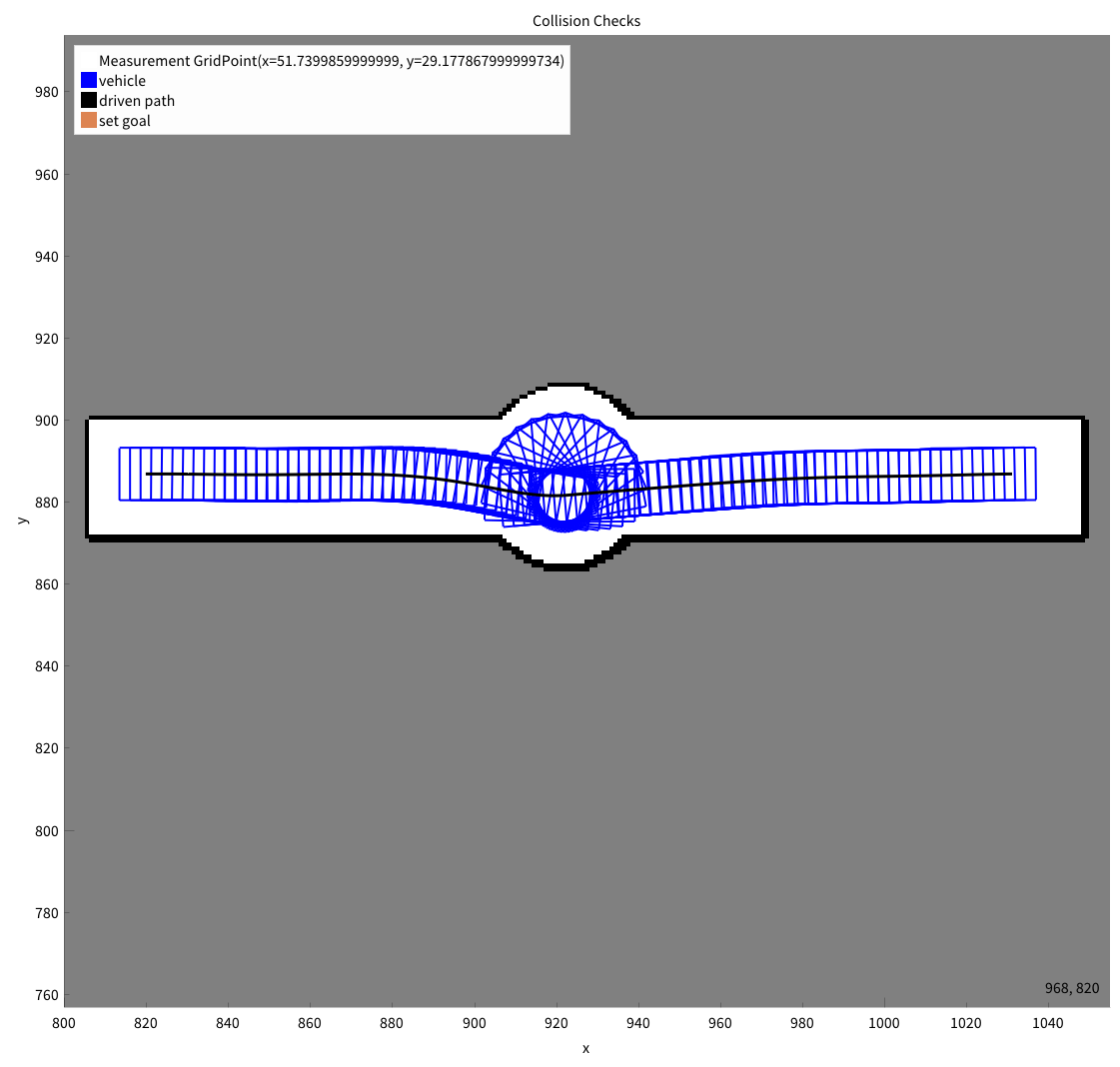}%
        \end{minipage}\\
\raggedright
\begin{minipage}{0.925\columnwidth}
\raggedleft
\begin{tikzpicture}
\begin{axis}[axis line style={draw=none}, tick style={draw=none}, yticklabels={,,}, xticklabels={,,}, width=3cm, height=1.7cm, legend columns=3]
\addplot[color=red, forget plot]{exp(x)};
\addlegendimage{area legend, red}
\addlegendentry{standard HA*}
\addlegendimage{area legend, blue}
\addlegendentry{extended HA*}
\addlegendimage{black}
\addlegendentry{path}
\end{axis}
\end{tikzpicture}
\end{minipage}
	\caption{Path comparison in two narrow environments. a) and c) show the narrow environment, b) and d) the even narrower variant.}
\label{fig:ra_qualitative}
\end{figure}
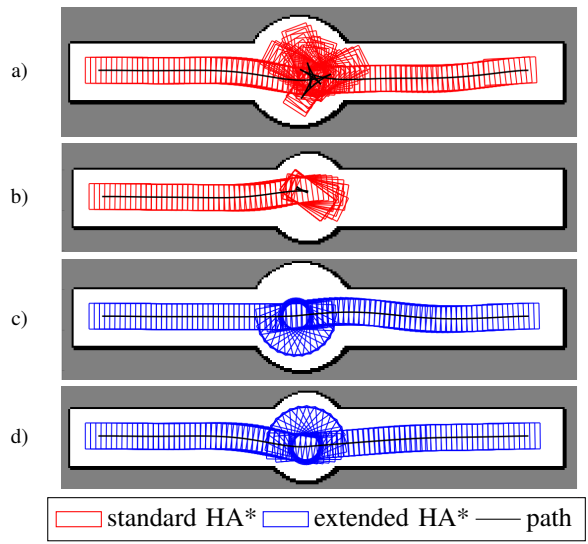

Next, it is investigated, if the U-Shift vehicle can benefit from the extended version of the path planner to improve its maneuvering capabilities in narrow scenarios. Up to the date of publication, the driveboard is still in construction. Hence, the extended Hybrid A* algorithm can only be investigated in simulations.
Fig.~\ref{fig:ra_qualitative} compares the driven paths of the standard and extended Hybrid A* algorithm in two narrow environments. Here, a narrow corridor with a circular plate is shown. The circular plate has a diameter of \qty{8.4}{m} and \qty{6.7}{m} respectively. The goal is to reach the pose at the end of the corridor with opposite orientation. The simulated U-Shift has a length of $l=\qty{4}{m}$ and a width of $w=\qty{2}{m}$. It can be observed that in the first environment, the goal pose can be reached by the standard and the extended algorithm. However, the standard planner must switch directions seven times. In the narrower environment, the goal cannot be reached by the standard algorithm. Here, the extended algorithm can reach the goal pose similarly as in the first case.
\subsubsection{Simulation Results}
\begin{figure}[bp]
	     \begin{minipage}{0.15\columnwidth}
     \centering
     \footnotesize{a)}
     \end{minipage}
    \begin{minipage}{0.7\columnwidth}
        \frame{\includegraphics[trim={3cm 5.5cm 5.5cm 16cm},clip, width=1.0\columnwidth]{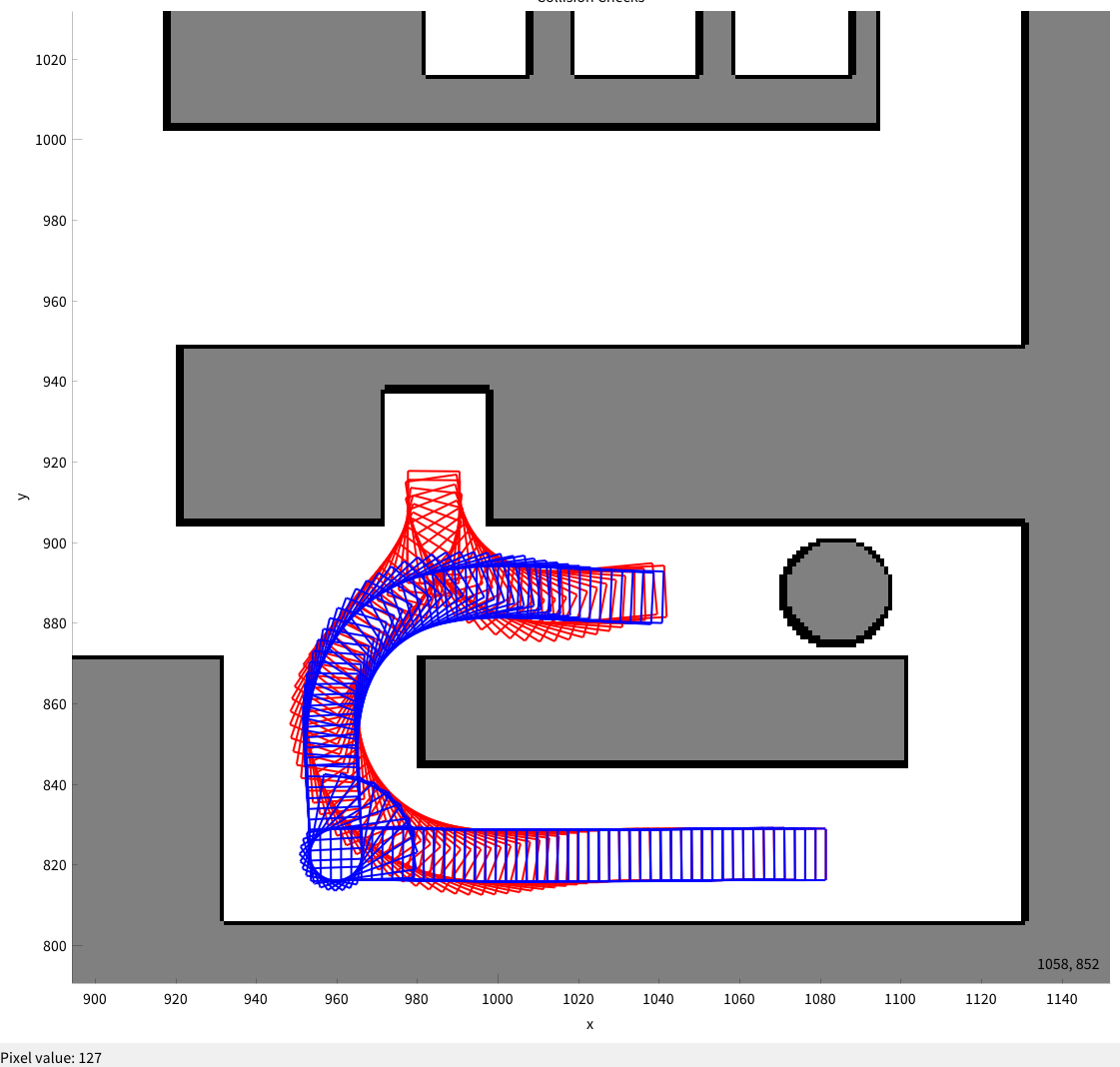}}
     \end{minipage}\\
    \vspace{1mm}\\
     \begin{minipage}{0.15\columnwidth}
      \centering
     \footnotesize{b)}
     \end{minipage}
     \begin{minipage}{0.7\columnwidth}
	\frame{\includegraphics[trim={8.5cm 10.5cm 0.5cm 12cm},clip, width=1.0\columnwidth]{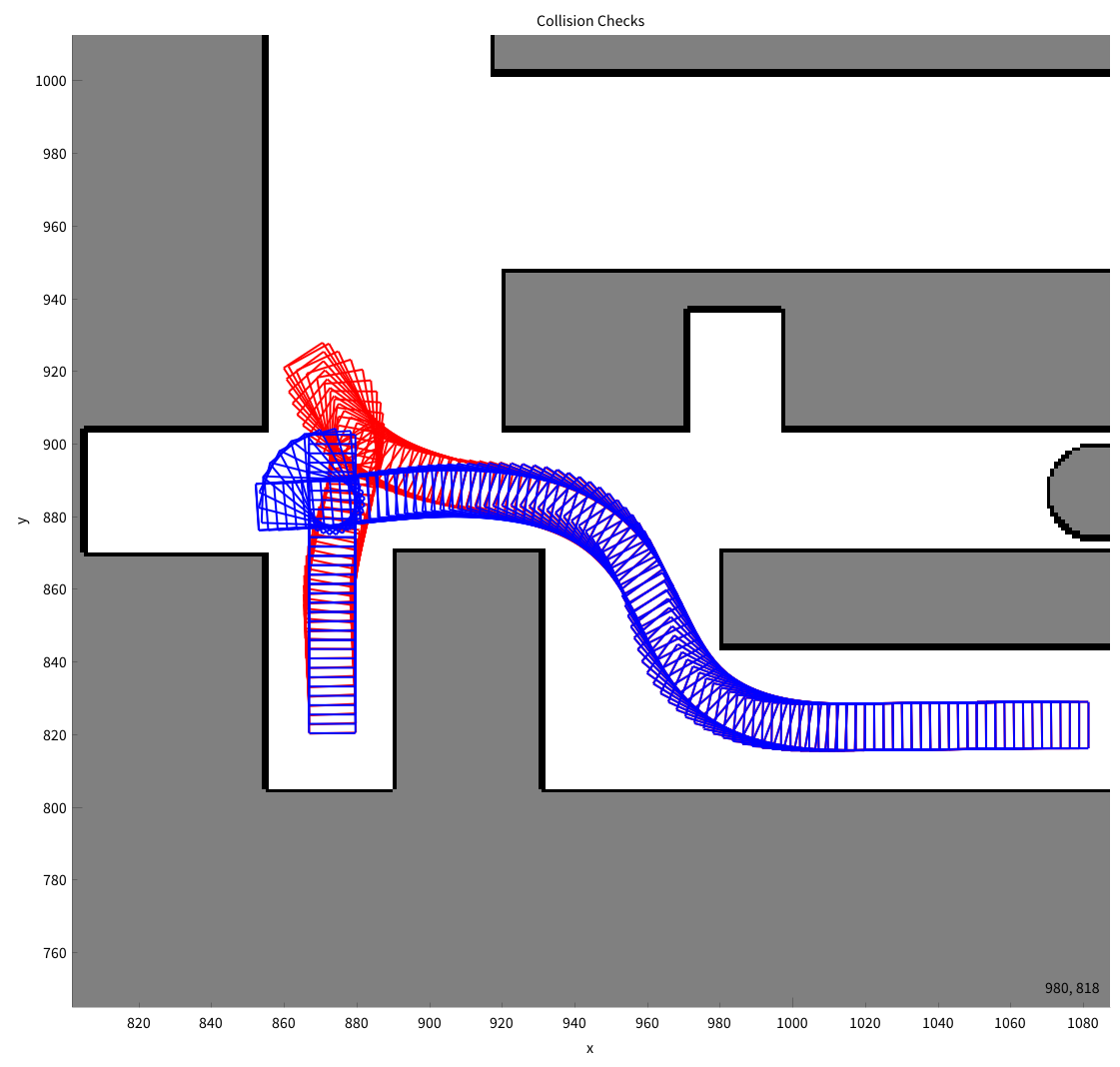}}%
    \end{minipage}\\
\raggedright
\begin{minipage}{0.85\columnwidth}
\raggedleft
\begin{tikzpicture}
\begin{axis}[axis line style={draw=none}, tick style={draw=none}, yticklabels={,,}, xticklabels={,,}, width=3cm, height=1.7cm, legend columns=3]
\addplot[color=red, forget plot]{exp(x)};
\addlegendimage{area legend, red}
\addlegendentry{standard HA*}
\addlegendimage{area legend, blue}
\addlegendentry{extended HA*}
\end{axis}
\end{tikzpicture}
\end{minipage}
	\caption{a) Use of the additional motion primitives. b) Use of the geometric extension.}
	\label{fig:nice_ra}
\end{figure}
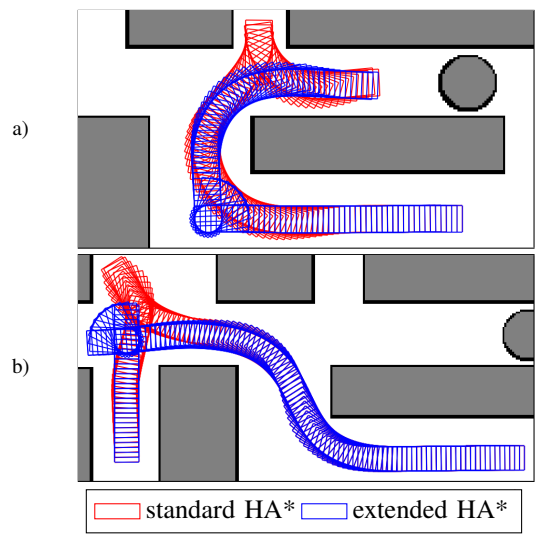%
Now, the standard algorithm is compared against the extended version in two realistic scenarios that are shown in Fig.~\ref{fig:nice_ra}. The corresponding videos generating the given paths can be found here \footnote{\href{https://youtu.be/5uQWnyPqYFw}{https://youtu.be/5uQWnyPqYFw}}. In the first scenario, the use of additional motion primitives during the graph exploration can be observed. The vehicle turns around early on the rear axis to approach the goal driving forward without the need to change directions in the narrow passage at the top. In the second scenario, the application of the rear axis extension is shown. It decreases the needed area for the direction switch which leads to a more concise path.

\begin{table}[tbp]
\vspace{2mm}
	\caption{Quantitative comparison of standard and extended Hybrid~A*}
	\label{tab:quant_extended}
	\begin{center}
		\begin{tabular}{c  c | c  c | c  c | c  c}
			\toprule
			Scen. & Method & $\dot{\kappa}\textsubscript{RMS}$ & $\mid \dot{\kappa}\textsubscript{max} \mid$ & $p\textsubscript{max}$ & $p\textsubscript{avg}$ & $l$ & $t\textsubscript{avg}$ \\
           & & \multicolumn{2}{c|}{in $1/\si{\meter}^2$} &\multicolumn{2}{c|}{-} & in \si{\meter} & in \si{\second}\\
			\midrule
			\multirow{2}{*}{(1)} & standard  
			& 1.28
			& 15.7
			& 1.0
			& 0.04
			& 37
			& \textbf{0.04}
			\\
			& extended
			& \textbf{0.86}
			& \textbf{10.5}
			& \textbf{0.45}
			& \textbf{0.02}
			& \textbf{34}
			& 0.05
			\\
            \midrule
			\multirow{2}{*}{(2)} & standard
			& 0.57
			& 9.0
			& 1.0
			& 0.03
			& 50
			& 0.07
			\\
			& extended & 
			\textbf{0.23}
			& \textbf{5.2}
			& 1.0
			& 0.03
			& \textbf{46}
			& 0.07
			\\
			\bottomrule
		\end{tabular}
	\end{center}
\end{table}%
Next, the mentioned metrics of smoothness, clearance, and length of paths are compared in Table~\ref{tab:quant_extended}.
The quantitative results show that the paths of the extended version are shorter and have an improved or equal proximity to objects. This shows that this capability can lead to safer and more concise paths. Furthermore, the extended method has an improved mean and maximum $\dot{\kappa}$, which was predictable as most of the relevant orientation change is done during the rotation. Here, $\kappa$ cannot be calculated as the vehicle does not move any distance $ds$ between two points. At last, the execution times of the scenarios are nearly equal, which confirms the measured runtimes in Table~\ref{tab:extended_versus_vanilla}.

To summarize, the proposed path planner uses the maneuvering capabilities of the U-Shift~II concept vehicle inherently which can be immensely helpful in narrow environments.

\section{CONCLUSIONS} \label{sec:conclusion}
In this paper, we illustrated methods to efficiently plan paths in large and unexplored environments. We stated the problem of using two separate algorithms for navigation and planning of existing navigation methods and developed a novel method called \textit{early stopping} that solves these problems inherently leading to reduced planning times with improved paths with respect to the system model. This method was applied to real-world data and investigated in simulations. It was shown that the guided path planning reduces the average and cumulative runtime by a factor of $\approx4.6$ and $\approx2$ respectively in our simulations while retaining smoothness, clearance, and length of the path. Further, we proposed an extension to the Hybrid A* algorithm that enables us to plan concise paths in narrow environments for vehicles with switchable system models as the U-Shift~II concept vehicle. In future work, one could adapt this planning approach to the UNICAR\textit{agil} vehicle. Furthermore, one can investigate to what extent the guided and extended methods can be adapted to sampling-based path planning approaches.

\addtolength{\textheight}{-12cm}   



%



\bibliographystyle{./IEEEtran.bst} 
\bibliography{./IEEEabrv,./itsc2023}

\end{document}